\def\year{2021}\relax
\documentclass[letterpaper]{article} 
\usepackage{aaai21}  
\usepackage{times}  
\usepackage{helvet} 
\usepackage{courier}  
\usepackage[hyphens]{url}  
\usepackage{graphicx} 
\usepackage{natbib}  
\usepackage{caption} 

\usepackage{amsmath,amssymb}
\usepackage{multirow}
\usepackage{color}

\usepackage[switch]{lineno}

\title{Semantic Editing On Segmentation Map Via Multi-Expansion Loss}
\author{

    Jianfeng He{$^1$}, Xuchao Zhang{$^1$}, Shuo Lei{$^1$},
Shuhui Wang{$^2$},\\
\textbf{Qingming Huang{$^{2, 3}$}, Chang-Tien Lu{$^1$}, Bei Xiao{$^4$}}\\ 
 {$^1$}Discovery Analytics Center, Virginia Tech, Falls Church, VA, USA\\
   {$^2$}Key Laboratory of Intelligent Information Processing of Chinese Academy of Sciences (CAS),
Institute of Computing Technology, CAS, Beijing, China\\
{$^3$}University of Chinese Academy of Sciences, Beijing, China\\
{$^4$}Department of Computer Science, American University, Washington, DC, USA\\
 {$^1$}\{jianfenghe, xuczhang, slei, ctlu\}@vt.edu, 
 \\{$^2$}wangshuhui@ict.ac.cn, {$^3$}qmhuang@ucas.ac.cn, {$^4$}bei.xiao@american.edu
}
\affiliations{

}

\begin{document}


\maketitle

\begin{abstract}

Semantic editing on segmentation map has been proposed as an intermediate interface for image generation, because it provides flexible and strong assistance in various image generation tasks. This paper aims to improve quality of edited segmentation map conditioned on semantic inputs. Even though recent studies apply global and local adversarial losses extensively to generate images for higher image quality, we find that they suffer from the misalignment of the boundary area in the mask area. To address this, we propose MExGAN for semantic editing on segmentation map, which uses a novel Multi-Expansion (MEx) loss implemented by adversarial losses on MEx areas. Each MEx area has the mask area of the generation as the majority and the boundary of original context as the minority. To boost convenience and stability of MEx loss, we further propose an Approximated MEx (A-MEx) loss. Besides, in contrast to previous model that builds training data for semantic editing on segmentation map with part of the whole image, which leads to model performance degradation, MExGAN applies the whole image to build the training data. Extensive experiments on semantic editing on segmentation map and natural image inpainting show competitive results on four datasets. 

\end{abstract}

\section{Introduction}
\label{sec:intro}

\begin{figure}[htb]
\centering
\includegraphics[width=0.45\textwidth]{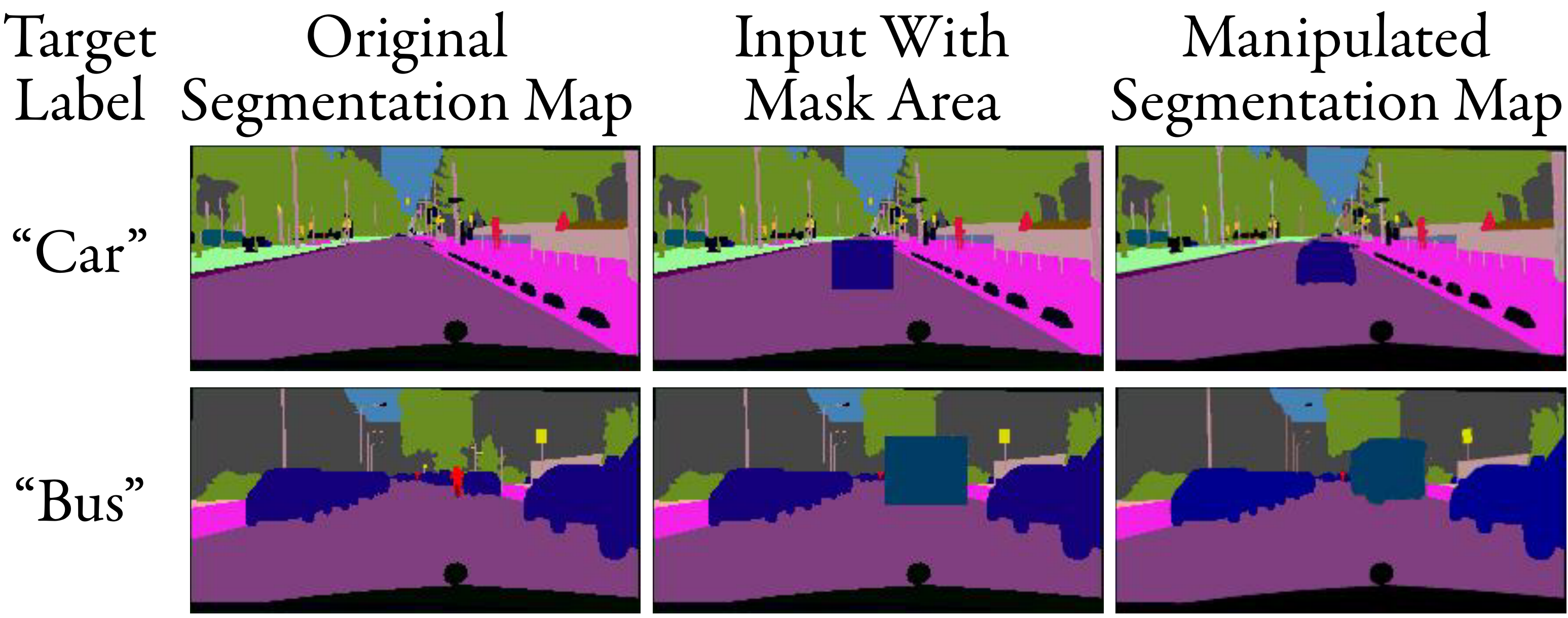}
\caption{Examples of semantic editing on segmentation map in Cityscape~\protect\cite{Cordts2016Cityscapes}. In the first row, we add a ``car" in the masked area. And in the second row, we heterogeneously replace a car and a person with a ``bus".}
\label{fig:example1}
\end{figure}



Deep learning has achieved impressive results in image generation such as image manipulation. Editing content directly on segmentation map has been recently proposed as an intermediate interface for image generation~\protect\cite{hong2018learning} due to its versatile abilities and great potential to improve performance in various generation tasks. 


This paper focuses on semantic editing on segmentation map, shown as Fig.~\ref{fig:example1}. It performs the editing automatically, rather than manually, such as manual drawing by users.
Plus, it generates manipulated results reflecting semantics defined by a target label, rather than only being consistent with the context, such as image inpainting~\protect\cite{yi2020contextual}. Our work improves the performance of semantic editing on segmentation map by solving a main challenge, explained below.

\begin{figure}[htbp]
\centering
\includegraphics[width=0.45\textwidth]{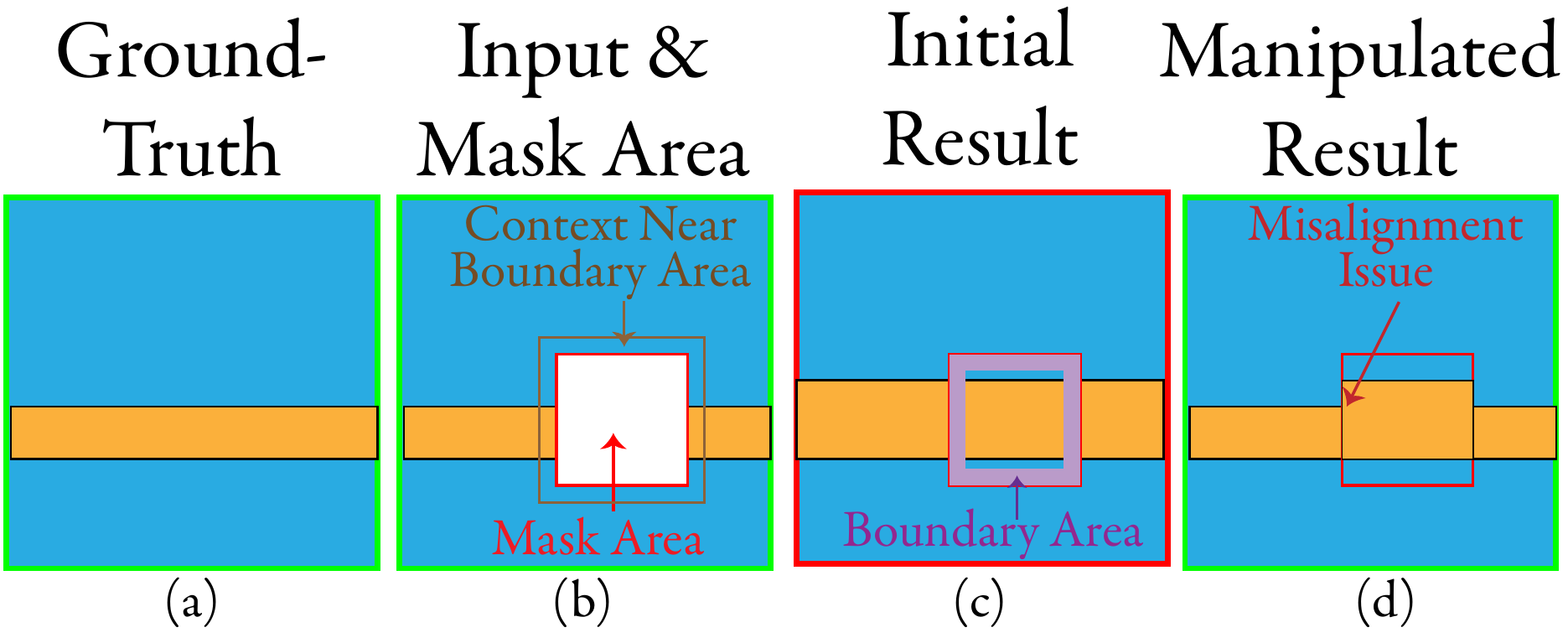}
\caption{Given a ground-truth (a), the misalignment issue in manipulated result (d) is caused by misalignment between the initial result (c) and original context from (b). Green border shows original context, red one shows generated context.}
\label{fig:seamless}
\end{figure}

\textbf{Challenge} The current applications of global and local adversarial losses~\protect\cite{iizuka2017globally,wang2018image}, which aim to improve quality of generated images, tend to result in misalignment of the boundary area in the mask area. Previous works show applying both global and local losses achieves better results, in comparison with applying the global loss alone. We conjecture this is caused by the inability to penalize all fake parts by the global adversarial loss alone, due to the small percentage of pixels composing image fine details. 
On the other hand, the local adversarial loss is too weak to boost the perceptual quality in \textit{areas near the boundary of the mask area} (abbreviated as Boundary Area, shown in Fig.~\ref{fig:seamless}(c)), because it has no knowledge of the context near Boundary Area (see Fig.~\ref{fig:seamless}(b)) in the original context. 
These drawbacks cause the \textit{misalignment of the Boundary Area} (abbreviated as misalignment issue, shown in Fig.~\ref{fig:seamless}(d)). Explicitly, even if we apply both the global and local adversarial losses, the Boundary Area might still be a blind spot. Thus, the Boundary Area in the initial generated result (abbreviated as initial result, shown as Fig.~\ref{fig:seamless}(c)) might cause misalignment with the ground-truth (shown as Fig.~\ref{fig:seamless}(a)). Additionally, our task fuses the mask area (shown as Fig.~\ref{fig:seamless}(b)) of the initial result with the original context to obtain the manipulated result (Fig.~\ref{fig:seamless}(d)). Together, these cause artifacts in Boundary Area of the manipulated result.  

We also improve data preparation used in the state-of-the-art model of our task. 
Because human annotation is difficult, ~\protect\cite{hong2018learning} proposes self-supervised learning to prepare the training data. However, they only use part of the whole image as a conditional input, which causes information loss due to the removed context, the difference between the chosen area and the full image. We believe the removed context includes the structure and semantic information that can guide the generation of mask area.


 
Thus, we propose Multi-Expansion GAN (MExGAN). MExGAN applies full images as a conditional input to improve data preparation. It can also  flexibly replace basic generators for other applications, such as in natural image inpainting. To tackle the misalignment issue, MExGAN applies a novel Multi-Expansion (MEx) loss using adversarial losses on the MEx areas. Specifically, MExGAN constructs manipulated segmentation map by fusing the initial result and original context. Then, it chooses multiple MEx areas by expanding mask areas multiple times in the manipulated segmentation map. Due to the expansion operations, each MEx area has mask area from the initial result as a principal component and the partially original context as an assisting component. To boost convenience and stability of MEx loss, we further propose an Approximated MEx (A-MEx) loss.


We set interactive image translation~\protect\cite{wang2018high} as a downstream task, which translates the manipulated segmentation maps into natural images. This is different from the semantic image inpainting used in~\protect\cite{hong2018learning}. The results demonstrate diverse abilities of our model.

To summarize, the following are our main contributions:
\begin{itemize}
  \item \textbf{Improvement and flexibility in MExGAN.} We propose a framework MExGAN for our task, which improves data preparation using the whole images directly. MExGAN can also flexibly replace basic structure generator in other applications such as image inpainting.

\item \textbf{Propose MEx loss.} To address the misalignment issue in image generation with global and local adversarial losses, we propose MEx loss, implemented by adversarial losses on MEx areas. Each MEx area has the mask area of initial result as the majority part and the partial original context as the minority part. 

\item \textbf{Propose A-MEx losses.} An approximation of MEx loss, A-MEx Loss, is proposed to boost convenience and stability of MEx loss.
  

  \item \textbf{Extensive experiments on two tasks.} Extensive experiments are conducted on four datasets. We show competitive results and demonstrate the beneficial effects of MEx loss on two tasks, semantic editing on segmentation map and image inpainting on natural images. 
  
  

 
\end{itemize}

\section{Related Work}
\label{sec:related}
\textbf{Image manipulation} has made tremendous progress by recent rapid development of GAN~\protect\cite{goodfellow2014generative}, and Variational Auto Encoder (VAE)~\protect\cite{kingma2013auto}. Based on the types of input domain, these methods can be divided into four categories, {\it i.e.}, directly on natural images~\protect\cite{zhu2016generative,yu2018generative,zhu2017unpaired}, indirectly on sketches~\protect\cite{chen2018sketchygan,qiu2019semanticadv}, segmentation maps~\protect\cite{wang2018high,isola2017image,chen2017photographic,park2019semantic}, and scene graphs~\protect\cite{johnson2018image,mittal2019interactive}. 
In terms of the tasks and functions, it can be divided into two categories, manipulation on content, such as adding~\protect\cite{yu2018generative,wang2018high,yao20183d} or removing objects~\protect\cite{shetty2018adversarial}; and manipulation on attributes~\protect\cite{mo2018instagan,park2019semantic,qiu2019semanticadv,yao20183d}, such as altering styles~\protect\cite{wang2016generative}. Our work focuses on semantically editing content of the segmentation maps.

\textbf{Segmentation map editing for image manipulation} has recently been applied. For examples,~\protect\cite{ding2019context}, ~\protect\cite{ntavelis2020sesame} and~\protect\cite{hong2018learning} generate the segmentation maps of target objects, which are later translated to natural images for semantic image editing. Similar to the framework used in above studies,~\protect\cite{liao2020guidance} and ~\protect\cite{song2018spg} focus on image inpainting. And~\protect\cite{mo2018instagan} transfers object shapes in the segmentation maps for attribute editing in natural images.

\textbf{Difference between MEx Loss and other common techniques} is obvious.
Attention mechanism, generally, learns weighted scores in pixel-level~\protect\cite{zhou2020learning} or feature-level~\protect\cite{zhou2020end}, so that the model can have different attention in the inputs. Partial convolution~\protect\cite{liu2018image} filters the information from mask areas in the convolution operation to reduce noises from mask areas. Multi-scale resizes the full images into various sizes for hierarchical feature representation~\protect\cite{wang2020self}.
In contrast, MEx loss focuses on solving the misalignment issue in Boundary Area using MEx areas. Also, MEx loss is dynamic through changing the radius for each expansion, rather than static, such as conventional attention mechanism and partial convolution. Importantly, MEx loss provides multiple mid-level views, which are ignored in global and local views. For example, in multi-scale, it still focuses on the global view.

\section{Model}
\label{sec:model}
Given a complete segmentation map $\mathop{ \mathbf{S}}^{c} \in \mathbb{R}^{H\times W \times 1}$ and its colorful version $\mathop{ \mathbf{Y}}^{c} \in \mathbb{R}^{H\times W \times 3}$, where $\mathop{ \mathbf{S}}$, $\mathop{ \mathbf{Y}}$, $H$ and $W$ represent the single-channel segmentation map, RGB-channel segmentation map, height and width respectively. We aim to synthesize the manipulated segmentation map $\mathop{ \mathbf{\hat{Y}}} \in \mathbb{R}^{H\times W \times 3}$ with semantics defined by a target label $l^t$. 

We choose an object bounding box $B$, which guides us to generate $\mathop{ \mathbf{A}}^{u}$ by the area of $B$ on $\mathop{ \mathbf{A}}^{c}$, where $\mathop{ \mathbf{A}}$ can be $\mathop{ \mathbf{S}}$ or $\mathop{ \mathbf{Y}}$.
Specifically, we define an object bounding box $B=\{\mathbf{b}, l^t\}$, as a combination of box corner $\mathbf{b}\in \mathbb{R}^4$ and a target label $l^t$. A user can semantically edit a segmentation map by setting an arbitrary bounding box and a target label. 

\begin{figure*}[htb]
\centering
\includegraphics[width = 0.95\textwidth]{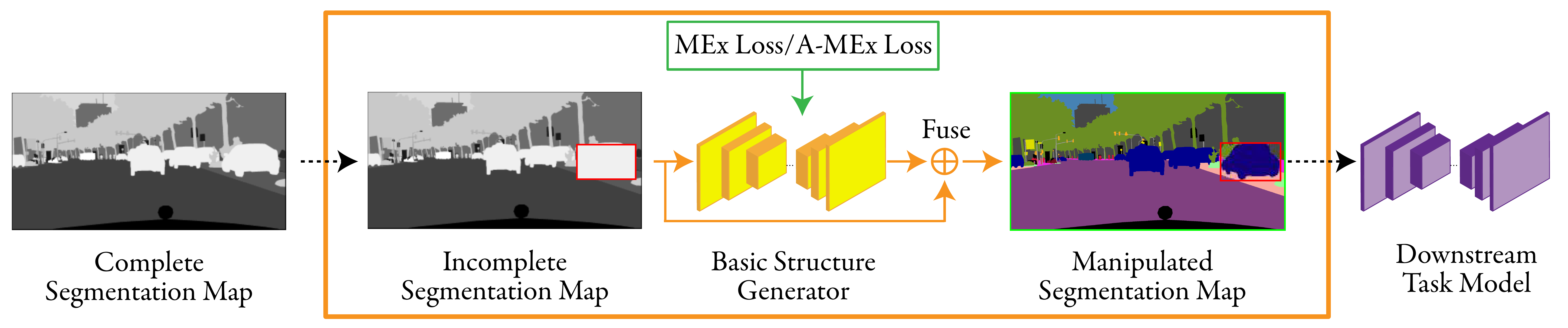}
\caption{Diagram of the overall framework for MExGAN. With complete segmentation map $\mathop{ \mathbf{S}}^{c}$, we can mask its all pixels in an object bounding box $B=\{\mathbf{b}, l^t\}$ as $l^t$ to build an incomplete segmentation map $\mathop{ \mathbf{S}}^{u}$. Conditioned on ($\mathop{ \mathbf{S}}^{u}$, $B$) and supervised by complete segmentation map $\mathop{ \mathbf{Y}}^{c}$ in color, the basic structure generator outputs initial segmentation map $ \mathbf{\widetilde{Y}}$, which is fused with $\mathop{ \mathbf{S}}^{u}$ to obtain manipulated segmentation map $ \mathbf{\hat{Y}}$. Finally, the $ \mathbf{\hat{Y}}$ is sent to a downstream task model.
With MEx Loss (shown as Fig.~\ref{fig:example3}) or A-MEx loss, the basic structure generator becomes MExGAN. The $\mathop{ \mathbf{Y}}^{c}$ and $ \mathbf{\widetilde{Y}}$ are not drawn in the figure.}
\label{fig:example2}
\end{figure*}

Fig.~\ref{fig:example2} shows the overall pipeline for MExGAN. First, MExGAN constructs an incomplete segmentation map $\mathop{ \mathbf{S}}^{u} \in \mathbb{R}^{H\times W \times 1}$ by copying $\mathop{ \mathbf{S}}^{c}$ and masking all pixels in the bounding box $B$ as $l^t$, which informs the model about the location and semantics to generate. It is important to point out that the way we use $\mathop{ \mathbf{S}}^{c}$ is different from~\protect\cite{hong2018learning}: we apply the global observation of $\mathop{ \mathbf{S}}^{c}$, while~\protect\cite{hong2018learning} applies only a local squared observation of $\mathop{ \mathbf{S}}^{c}$, which omits much of the removed context that belongs to the same image. Removing the contexts will therefore reduce the quality of the generated results. Second, MExGAN generates its initial segmentation map $\mathop{ \mathbf{\widetilde{Y}}}$ by $\mathop{ \mathbf{\widetilde{Y}}}=G^{Y}(\mathop{ \mathbf{S}}^{u}, B)$, conditioned on $\mathop{ \mathbf{S}}^{u}$ and $B$. Specifically, $G^Y$ is a learnable structure generator and $\mathop{ \mathbf{\widetilde{Y}}}$ is supervised by $\mathop{ \mathbf{Y}}^{c}$. Next, we fuse the structure generator output $\mathop{ \mathbf{\widetilde{Y}}}$ and $\mathop{ \mathbf{Y}}^{u}$ to derive the manipulated segmentation map $\mathop{ \mathbf{\hat{Y}}}$.
Finally, $\mathop{ \mathbf{\hat{Y}}}$ is sent to a downstream model to perform a subsequent specific task such as generating a natural image. In this paper, to demonstrate the flexibility of our pipeline, we apply image translation as our downstream task, rather than semantic image inpainting shown in ~\protect\cite{hong2018learning}.

In the next subsections, we begin by introducing a basic structure generator $G^{0}$ without the MEx loss. We then describe MEx loss and add the loss to the basic structure generator, so that we have MEx structure generator $G^{Y}$. We also propose an Approximated MEx (A-MEx) loss to boost convenience and stability of MEx loss. We end by briefly describing a pre-trained downstream task model $G^{I}$.

\subsection{Basic Structure Generator}
The goal of the basic structure generator $G^{0}$ is to infer the segmentation map of the masked region $\mathbf{M}$ in $\mathop{ \mathbf{S}}^{u}$, where $\mathbf{M}\in\mathbb{R}^{H\times W}$ is a binary matrix specified by $B=\{\mathbf{b}, l^t\}$: $M_{ij}=1$ for all pixels $(i, j)$ inside the bounding box $B$. The output $\mathop{ \mathbf{\widetilde{Y}}}$ of structure generator $G^{0}$, which is an initial output for our task, should reflect the class-specific structure of the object defined by $l^t$. The output $\mathop{ \mathbf{\widetilde{Y}}}$ should achieve consistency in context between generated object and its surrounded context. 
Considering the two requirements, the basic structure generator is conditioned on $\mathop{ \mathbf{S}}^{u}$ and $B$, so that we have $G^{0}$ by $\mathop{ \mathbf{\widetilde{Y}}}=G^{0}(\mathop{ \mathbf{S}}^{u}, B)$. 

Our design principles for the basic structure generator are motivated by an image translation model, Pix2PixHD~\protect\cite{wang2018high}, which translates segmentation maps to natural images. Based on the Pix2PixHD and our requirements, 
the objective function for our basic model is given by,
\begin{equation}
\begin{split}
\mathbf{\mathcal{L}}_{\mathrm{B}} = \lambda_1\mathbf{\mathcal{L}}_{\mathrm{adv}}({ \mathbf{\widetilde{Y}}},{ \mathbf{Y}}^{c}) + \\ \lambda_2\mathbf{\mathcal{L}}_{\mathrm{fea}}({ \mathbf{\widetilde{Y}}},{ \mathbf{Y}}^{c}) + \lambda_3\mathbf{\mathcal{L}}_{\mathrm{pec}}({ \mathbf{\widetilde{Y}}},{ \mathbf{Y}}^{c})
\label{eq:Basic}
\end{split}
\end{equation}
where the complete segmentation map $\mathop{ \mathbf{Y}}^{c}$ in color format is the ground-truth. ${\mathcal{L}}_{\mathrm{adv}}({ \mathbf{\widetilde{Y}}},{ \mathbf{Y}}^{c})$ is the conditional adversarial loss defined on $\mathop{ \mathbf{S}}^{u}$ and $\mathop{ \mathbf{Y}}^{c}$ ensuring the perceptual quality of predicted segmentation map, which is shown as,
\begin{equation}
\begin{split}
{\mathcal{L}}_{\mathrm{adv}}({ \mathbf{\widetilde{Y}}},{ \mathbf{Y}}^{c}) = \mathbb{E}_{{ \mathbf{Y}}^{c}}[log(D^{Y}({ \mathbf{Y}}^{c},{ \mathbf{S}}^{u}))] + \\
\mathbb{E}_{{ \mathbf{\widetilde{Y}}}}[1-log(D^{Y}({ \mathbf{\widetilde{Y}}},{ \mathbf{S}}^{u}))]
\label{eq:adv}
\end{split}
\end{equation}
where $D^Y$ is a learnable conditional discriminator. In Eq.~\ref{eq:Basic}, $\mathbf{\mathcal{L}}_{\mathrm{fea}}({ \mathbf{\widetilde{Y}}},{ \mathbf{Y}}^{c})$ is the feature matching loss~\protect\cite{wang2018high}, which computes the difference between the initial output and ground-truth using the intermediate feature from $D^Y$ by,
\begin{equation}
{\mathcal{L}}_{\mathrm{fea}}({ \mathbf{\widetilde{Y}}},{ \mathbf{Y}}^{c}) = \sum_{i=1}^{t}\frac{1}{N_i}||
D^{Y}_{i}({ \mathbf{Y}}^{c},{ \mathbf{S}}^{u}) - D^{Y}_{i}({ \mathbf{\widetilde{Y}}},{ \mathbf{S}}^{u})||_{1}
\label{eq:fea}
\end{equation}
where $t$ is the total number of layers in $D^{Y}$ and $N_i$ denotes the number of elements in each layer. The $\mathbf{\mathcal{L}}_{\mathrm{pec}}({ \mathbf{\widetilde{Y}}},{ \mathbf{Y}}^{c})$ in Eq.~\ref{eq:Basic} is the perception loss, which can further boost the perceptual quality of the initial output by,
\begin{equation}
{\mathcal{L}}_{\mathrm{pec}}({ \mathbf{\widetilde{Y}}},{ \mathbf{Y}}^{c}) = \sum_{i=1}^{r}w_{i}||
E_{i}( \mathbf{\widetilde{Y}})-E_{i}({ \mathbf{Y}}^{c})||_{1}
\label{eq:per}
\end{equation}
where $r$ is the number of chosen layers, $w_i$ is a pre-set weight for difference from each chosen layer, $E_i$ is $i$-th chosen layer in a pre-trained encoder. We set $r=5$ and $E$ as VGG19~\protect\cite{simonyan2014very} in our experiment.

With loss function defined in Eq.~\ref{eq:Basic}, we can train $G^{0}$. During the inference, we construct the manipulated segmentation map $ \mathbf{\hat{Y}}$ by fusing $ \mathbf{\widetilde{Y}}$ and $ \mathbf{Y}^{u}$. This is given by,
\begin{equation}
 \mathbf{\hat{Y}} =  \mathbf{\widetilde{Y}} \times \mathbf{M} +  \mathbf{Y}^{u} \times (\mathbf{1} - \mathbf{M})
\label{eq:fuse}
\end{equation}
where $\mathbf{1} \in \mathbb{R}^{H \times W}$ is a matrix with all elements setting to 1. Based on Eq.~\ref{eq:fuse}, we have $ \mathbf{\hat{Y}}$, which only replaces the mask area with the respective area in $ \mathbf{\widetilde{Y}}$, but keeps the rest the same as $ \mathbf{Y}^{u}$.

\begin{figure}[htb]
\centering
\includegraphics[width = 0.45\textwidth]{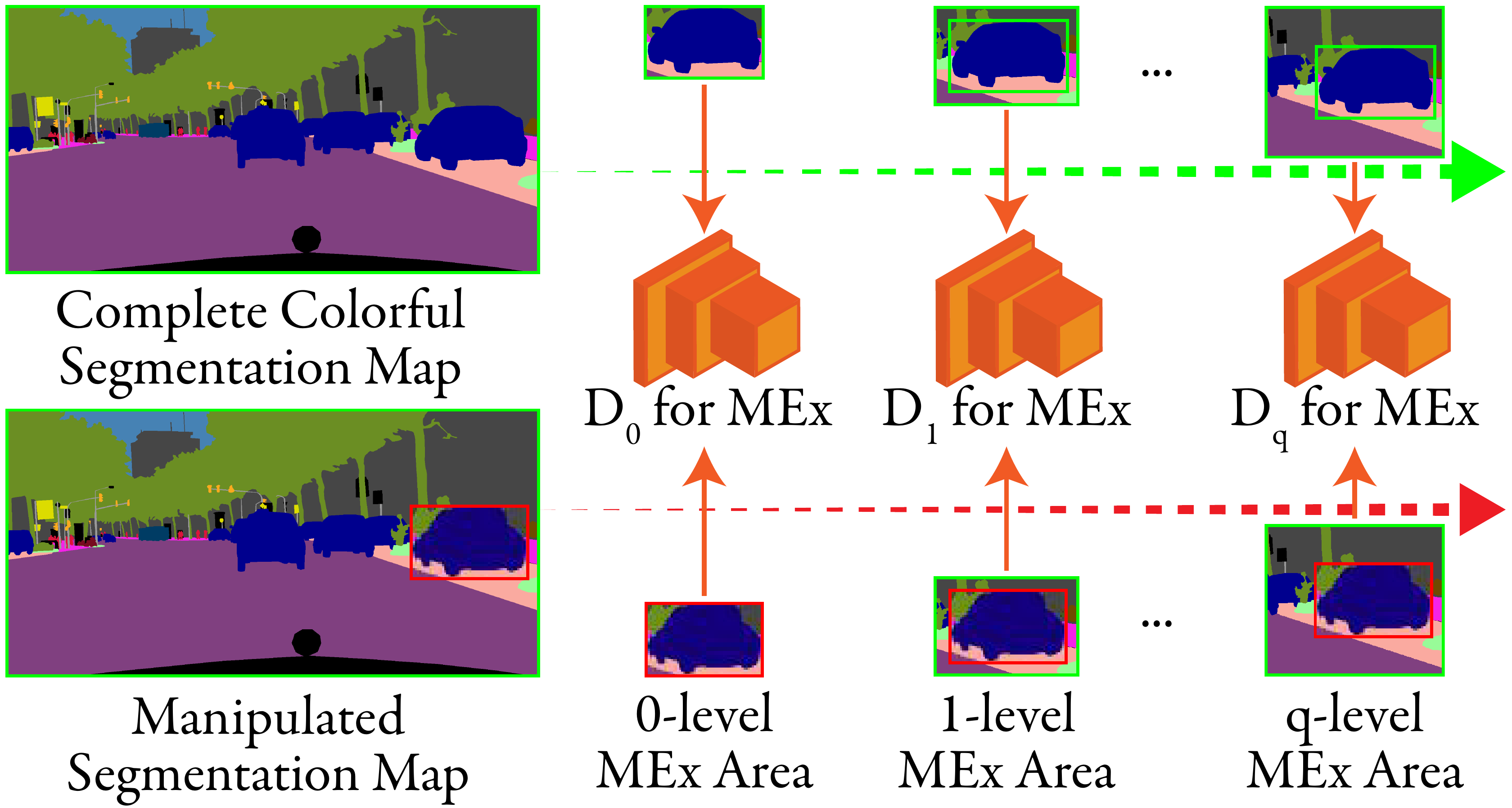}
\caption{Diagram of MEx loss. Given a complete colorful segmentation map $\mathop{ \mathbf{Y}}^{c}$, the green dotted arrow shows MEx areas $ \mathbf{{Z}}^{c}$ for ground-truth from the 0-th to the $j$-th level. Given a manipulated segmentation map $ \mathbf{\hat{Y}}$, the red dotted arrow represents a process to obtain MEx areas $ \mathbf{\hat{Z}}$ for the manipulated segmentation map. The green border and red border show ground-truth and generated area respectively.}
\label{fig:example3}
\end{figure}

\subsection{Multi-Expansion (MEx) Loss}
\label{sec:multiloss}
The improvement by adding extra local adversarial loss in~\protect\cite{iizuka2017globally} indicates that \textit{the global adversarial loss does not penalize all the fake parts in the generation (T1)}. Since the local adversarial loss improves the image quality and its input does not include context beyond the mask area, we assume that \textit{the drawback of global adversarial loss is caused by the huge difference in terms of number of pixels between context and the mask area, so that the loss from the mask area is not salient (T2)}. 

As for the losses described in Eq.~\ref{eq:Basic}, they are all designed for a global view ($\mathbf{\widetilde{Y}}$). 
Motivated by~\protect\cite{iizuka2017globally}, it is intuitive to consider local adversarial loss or other losses for a local view. The local view in our task is $( \mathbf{\widetilde{Y}} \times \mathbf{M})$. However, \textit{the local view does not include the boundary of context, which is the inner boundary of ${ \mathbf{Y}}^{u}$ in our task, because they have no intersection (T3)}. A drawback of having no knowledge about the boundary of context is that \textit{the losses on the local view is too weak to boost the realism in the Boundary Area (T4)}. 

Thus, based on \textit{(T1)} and \textit{(T4)}, \textit{the Boundary Area is more likely to be a blind spot for both adversarial losses, compared with the other areas in the mask area (T5)}.
Therefore, the perceptual quality of Boundary Area in the initial output $ \mathbf{\widetilde{Y}}$ is worse than other areas, even if we apply global and local adversarial losses together. Finally, based on Eq.~\ref{eq:fuse}, we can know that the Boundary Area has more chance to be misaligned with ${ \mathbf{Y}}^{u}$ (misalignment issue), due to \textit{(T5)}.

To solve the misalignment issue, we propose MEx loss, which calculates the adversarial losses for MEx areas. As shown in Fig.~\ref{fig:example3}, it is achieved by $(q+1)$ MEx discriminators, $D_0^E$, $D_1^E$, ..., $D_q^E$.
The key idea of the MEx loss is that we build MEx mask areas $\mathbf{M}^{E}\in\mathbb{R}^{H\times W}$, multiple expanded mask areas calculated by the mask area $\mathbf{M}$. Recall that $\mathbf{M}$ is constructed by box corners $\mathbf{b} = [b_1, b_2, b_3, b_4]$ from $B$, where $(b_1, b_2)$ and $(b_3, b_4)$ are coordinates of the left-top corner and the right-bottom corner for $B$. Then, at the $j$-th level expansion, we have $j$-level MEx mask area $\mathbf{M}^{E}_{j}$ constructed by $\mathbf{b}^{E}_j = [b_1-j\times \alpha, b_2 - j\times \beta, b_3 + j\times \alpha, b_4 + j\times \beta]$, where $\alpha$ and $\beta$ are step lengths in vertical and horizontal directions respectively. If any coordinate in $\mathbf{b}^{E}_j$ is smaller than $0$ or greater than $H-1$ or $W-1$, it is set to $0$ or $H-1$ or $W-1$ respectively. Based on $\mathbf{M}^{E}_{j}$, we have $j$-level MEx area $ \mathbf{{Z}}^{c}_{j}$ for ground-truth and $ \mathbf{\hat{Z}}_{j}$ for generation as,
\begin{equation}
 \mathbf{{Z}}^{c}_{j} = {Crop}( \mathbf{{Y}}^{c}\times  \mathbf{M}^{E}_{j}),
 \mathbf{\hat{Z}}_{j} = {Crop}( \mathbf{\hat{Y}}\times  \mathbf{M}^{E}_{j})
\label{eq:MEx_area}
\end{equation}
where $Crop$ is an operator to crop the input and only keep the area with non-zero elements.

From Fig.~\ref{fig:example3} and  Eq.~\ref{eq:MEx_area}, we see that the $j$-level MEx area $ \mathbf{\hat{Z}}_{j}$ consists of two parts, a local view for generation $ \mathbf{\widetilde{Y}}$ and its context from the ground-truth. Different from \textit{(T2)}, the losses from the mask area are more salient, because the mask area in each MEx Area is dominated by setting $\alpha$ and $\beta$ as small values compared with $H$ and $W$. Moreover, though the mask area is dominating, $D^E$ has knowledge of context of ${ \mathbf{Y}}^{u}$ by setting $q$ greater than $0$ (when $q=0$, MEx Loss is equivalent to the local adversarial loss), which avoids \textit{(T3)}. Hence, with the MEx areas sent to $D^E$, the MEx loss is more likely and able to penalize fake parts in the Boundary Area of $ \mathbf{\widetilde{Y}}$. In this way, the misalignment issue is addressed. 

The MEx loss function can be formulated as,
\begin{equation}
\begin{split}
{\mathcal{L}}_{\mathrm{MEx}}( \mathbf{\hat{Z}}, \mathbf{{Z}}^{c}) = \sum_{j=0}^{q} \mathbb{E}_{{ \mathbf{Z}}^{c}}[log(D^{E}_{j}( \mathbf{{Z}}^{c}_{j}, Crop({ \mathbf{S}}^{u} \times \mathbf{M}^{E}_{j})))] + \\ \sum_{j=0}^{q} \mathbb{E}_{{ \mathbf{\hat{Z}}}}[1-log(D^{E}_{j}({ \mathbf{\hat{Z}}_{j}} , Crop({ \mathbf{S}}^{u} \times \mathbf{M}^{E}_{j})))]
\label{eq:MEx_adv}
\end{split}
\end{equation}
where the condition $({ \mathbf{S}}^{u} \times \mathbf{M}^{E}_{j})$ is also cropped.

Finally, we define our MExGAN loss function as,
\begin{equation}
\begin{aligned}
{\mathcal{L}}_{\mathrm{MG}} = \mathbf{\mathcal{L}}_{\mathrm{B}} + \lambda_4 {\mathcal{L}}_{\mathrm{MEx}}( \mathbf{\hat{Z}}, \mathbf{{Z}}^{c})
\label{eq:MG}
\end{aligned}
\end{equation}
which adds MEx loss to the basic structure generator (Eq.~\ref{eq:Basic}). 

\subsubsection{Approximated MEx (A-MEx) Loss}
Though we have proposed MEx loss, we aim to boost convenience and stability of MEx loss by proposing an approximation. Initially, different from~\protect\cite{iizuka2017globally}, we find the $Crop$ operation leads to inputs in various sizes, which might require special design of the $D^E$. Thus, in A-MEx loss, we remove all $Crop$ in Eq.~\ref{eq:MEx_area} and Eq.~\ref{eq:MEx_adv}. Then, all MEx Areas sent to $D^E$ are the same size as $H\times W \times \{C\}$, where $\{C\}$ is their original channel numbers. 
As a result of having same size inputs, the stability of MEx loss is improved. Though we can design $D^E$ simply by removing the $Crop$ operation, the memory consumption is increased by enlarged input sizes. In addition, we find that MEx loss consumes much memory when $q$ is huge. Thus, in A-MEx loss, we do not apply $(q+1)$ MEx discriminators $D^E$, but apply only one $D^E$ to save memory. 
Specifically, the A-MEx loss function can be formulated as,
\begin{equation}
\begin{aligned}
{\mathcal{L}}_{\mathrm{A-MEx}}( \mathbf{\hat{Z}}, \mathbf{{Z}}^{c}) = \sum_{j=0}^{q} \mathbb{E}_{{ \mathbf{Z}}^{c}}[log(D^{E}( \mathbf{{Z}}^{c}_{j}, ({ \mathbf{S}}^{u} \times \mathbf{M}^{E}_{j})))] + \\ \sum_{j=0}^{q} \mathbb{E}_{{ \mathbf{\hat{Z}}}}[1-log(D^{E}({ \mathbf{\hat{Z}}_{j}}, ({ \mathbf{S}}^{u} \times \mathbf{M}^{E}_{j})))]
\label{eq:A_MEx_adv}
\end{aligned}
\end{equation}
where the $ \mathbf{\hat{Z}}_{j}$ and the $ \mathbf{{Z}}^{c}_{j}$ are calculated by Eq.~\ref{eq:MEx_area} without $Crop$. Since the current inputs to $D^E$ still have mask area as the majority and original context as the minority, A-MEx loss also addresses the misalignment issue. For convenience and stability, we can use A-MEx loss in Eq.~\ref{eq:MG}.

\subsubsection{Multiple Mid-Level Views In MEx Loss} 
Besides improving image quality by solving misalignment issue, the improvement by MEx loss can be explained from another perspective. In short, MEx loss provides multiple mid-level views in image generation. Concretely, compared with~\protect\cite{iizuka2017globally}, which applies global and local views, the MEx loss provides multiple mid-level views by the MEx areas from the $1$-th level to the $q$-th level. 
Then, similar to ensemble learning~\protect\cite{sagi2018ensemble}, each view can be seen as a voter. 
MEx loss can improve the image quality by providing more voters via multiple mid-level views. 

\subsection{Downstream Task Model}
Though there can be many choices for the downstream task, we apply image translation as our task, which interactively translates the manipulated segmentation map $ \mathbf{\hat{Y}}$ into natural image $I$. As for image translation model, we apply the pre-trained model $G^{I}$ from Pix2pixHD~\protect\cite{wang2018high}. This can be formulated as $I = G^{I}( \mathbf{\hat{Y}})$.

\section{Experiments}
\subsection{Experiment Setup}
\subsubsection{Datasets}
We conduct experiments on street scenes using ~\textbf{Cityscape} \protect\cite{Cordts2016Cityscapes}, indoor scenes using \textbf{NYU V2 (NYU)}~\protect\cite{Silberman:ECCV12}, and face images using rectified \textbf{Helen Face} datasets~\protect\cite{lin2019face}~\protect\cite{le2012interactive}. Specifically, we apply 2975 training images and 500 testing images for Cityscapes; 1200 training images and 249 testing images for  NYU; and 1800 training images and 299 testing images for Helen Face. We choose 8 categories of objects 
(`person', `rider', `car', `truck', `bus', `train', `motorcycle', and `bicycle') to edit for Cityscapes from its 35 object categories. For NYU, which has 895 semantic labels, we choose $7$ categories for editing. The Helen Face dataset has 11 categories, from which, we choose $8$ categories. These categories are chosen based on the number of objects in each dataset. To demonstrate  generality of MEx loss in natural images, we also perform image inpainting on architecture images from \textbf{CMP Facade}~\protect\cite{tylevcek2013spatial} dataset, from which 598 images are used for training and 8 images are applied for testing. In the testing, each natural image will be randomly masked for 20 times.

\begin{table*}[htb]
\begin{center}
\begin{tabular}{l|llll|lll|lll}
\hline
\multirow{2}{*}{Methods} & \multicolumn{4}{c|}{Cityscapes} & \multicolumn{3}{c|}{NYU}                   & \multicolumn{3}{c}{Helen Face}                   \\
                         & tIOU  &  hamm &FID &HuEv & tIOU & hamm &FID & tIOU &  hamm &FID \\
\hline
Pix2PixHD                    & 0.7350    & 0.7483 & 139.47 &0.4704  & 0.4222    & 0.3763 & 138.16   & 0.2806   & 0.4753  & 64.91  \\
\hline
MExGAN           & \bf 0.7493     & \bf 0.7677    & \bf139.14 &\bf0.5296  & \bf0.4851 & \bf0.4683  &\bf 137.44  & \bf0.3721     & \bf0.5278  & \bf63.90  \\
\hline
\end{tabular}
\caption{The tIOU, hamm, and FID on three datasets. Specifically, we also show the HuEv on Cityscapes.}
\label{table:pixHD}
\end{center}
\end{table*}

\subsubsection{Evaluation Metrics}
For our task, we use four metrics for the semantics and perceptual quality of the edited results: (1) \textbf{Target Intersection-Over-Union (tIOU)}~\protect\cite{liu2019learning}, which is the ratio of overlapped area for target object to union area for target object; (2) \textbf{Hamming Distance (hamm)}~\protect\cite{mun2017comparison}, which is the ratio of the number of pixels, having the same values as ground-truth, to the number of total pixels; (3) \textbf{Frechet Inception Distance (FID)}~\protect\cite{heusel2017gans}, which computes the distribution difference between a group of natural images 
and a group of translated natural images from the manipulated segmentation maps by a pretrained model; 
(4) \textbf{Human Evaluation (HuEv)}, which is the result of a two-alternative forced choices experiment, where users compare two model outputs and choose the better one in comparison with ground-truth.                                                                                                                                                                                                                                                                                                                                                                                                                                                                                                             
We measure performance by computing the percentage the counts of chosen model over total number of samples in a dataset averaging across observers.

For natural image inpainting, we apply two extra commonly-used metrics besides FID and HuEv: (1) \textbf{Structural Similarity Index (SSIM)}~\protect\cite{wang2004image}, which evaluates the similarity of the ground-truth and the generation based on low-level statistics; (2) \textbf{L1 Distance (L1)}, which averages the reconstruction loss for each pixel.

\subsubsection{Baselines and Ablation Setting}
We compare our model with three baselines: (1) \textbf{Pix2PixHD}~\protect\cite{wang2018high}, which is our basic generator with adversarial loss, feature match loss and perception loss; (2) \textbf{Global and Local model (GL)}~\protect\cite{iizuka2017globally}, which applies global adversarial loss and local adversarial loss together; (3) \textbf{Two-Stream Model (TwoSM)}~\protect\cite{hong2018learning}, which applies a double-stream model. Concretely, it has two outputs, a foreground map and a background map. The foreground map is a binary object mask defining the object shape tightly bounded by $B$. The background map defines both the object shape and its surrounded context in $B$. Since their two maps are generated based on shared embedding layers, each generation process effects each other. In contrast, Pix2PixHD and GL are both single-stream models, which have only one output including the target object shape and its surrounded context. 

To demonstrate the effect of MEx loss, the ablation setting has been designed in our experiments. \textbf{MExGAN} is actually a combination of Pix2PixHD and MEx loss. Similarly, we have \textbf{A-MExGAN} as a combination of Pix2PixHD and A-MEx loss. We also combine GL and A-MEx loss as \textbf{GL+A-MEx} to show effect of A-MEx loss in natural image inpainting. We have improved the self-supervised learning of TwoSM by replacing its local observation with our global observation, which is represented by ~\textbf{TwoSM+Full}. 

\subsubsection{Implementation Details}
 We train MExGAN and baselines for 200 epochs by default. We apply Adam~\protect\cite{kingma2014adam} as our optimizer with initial learning rate as 0.0002 and momentum as 0.5, decaying the learning rate from the 100-th epoch to the last epoch ending with learning rate as 0. We set $\lambda_1 =\lambda_2 = \lambda_3=1$ for basic structure generator in MExGAN and Pix2PixHD; set $\lambda_4=1$ for MExGAN and A-MExGAN. In the training, the bounding box is randomly selected for each image at each epoch, while in the testing, we compare two methods by choosing the same bounding box for each image.
We set $q=4$ for both Cityscapes and Helen Face, while set $q=5$ for NYU. We set $\alpha=\beta=5$ for all datasets by default. For image inpainting, we set $\lambda_4=1$, $q=\alpha=\beta=4$ for GL+A-MEx, and train both GL and GL+A-MEx for 2000 epochs.

\begin{figure}[htb]
\centering
\includegraphics[width = 0.47\textwidth]{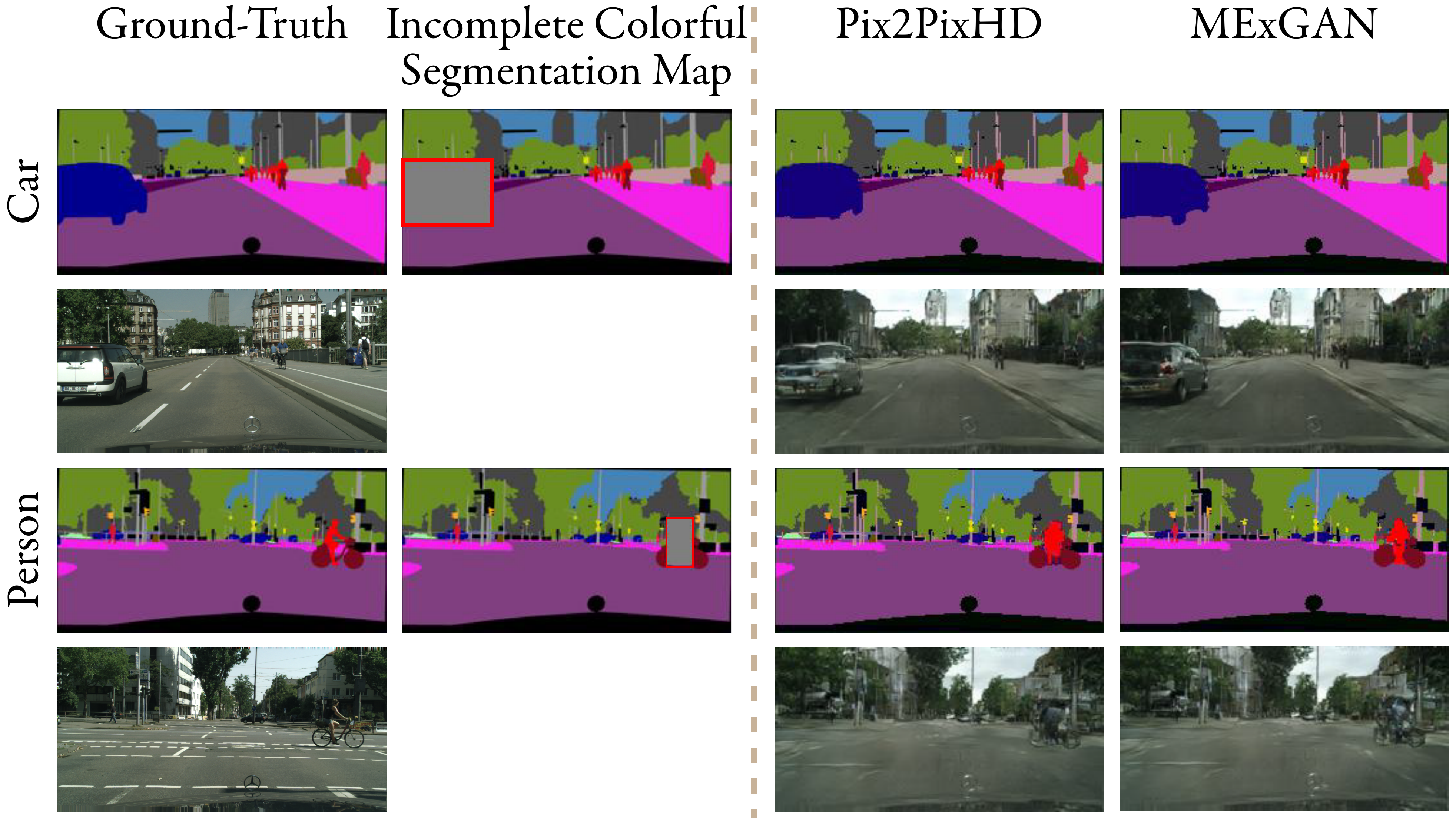}
\caption{Examples of manipulated segmentation maps of Pix2PixHD and MExGAN on the Cityscapes. The ground-truth (segmentation maps and natural images), and the incomplete color segmentation maps are shown on the left. The right two columns show the manipulated segmentation maps of Pix2PixHD and MExGAN respectively. The leftmost vertical texts are their target labels. }
\label{fig:Cityscapes}
\end{figure}

\begin{figure}[htb]
\centering
\includegraphics[width = 0.45\textwidth]{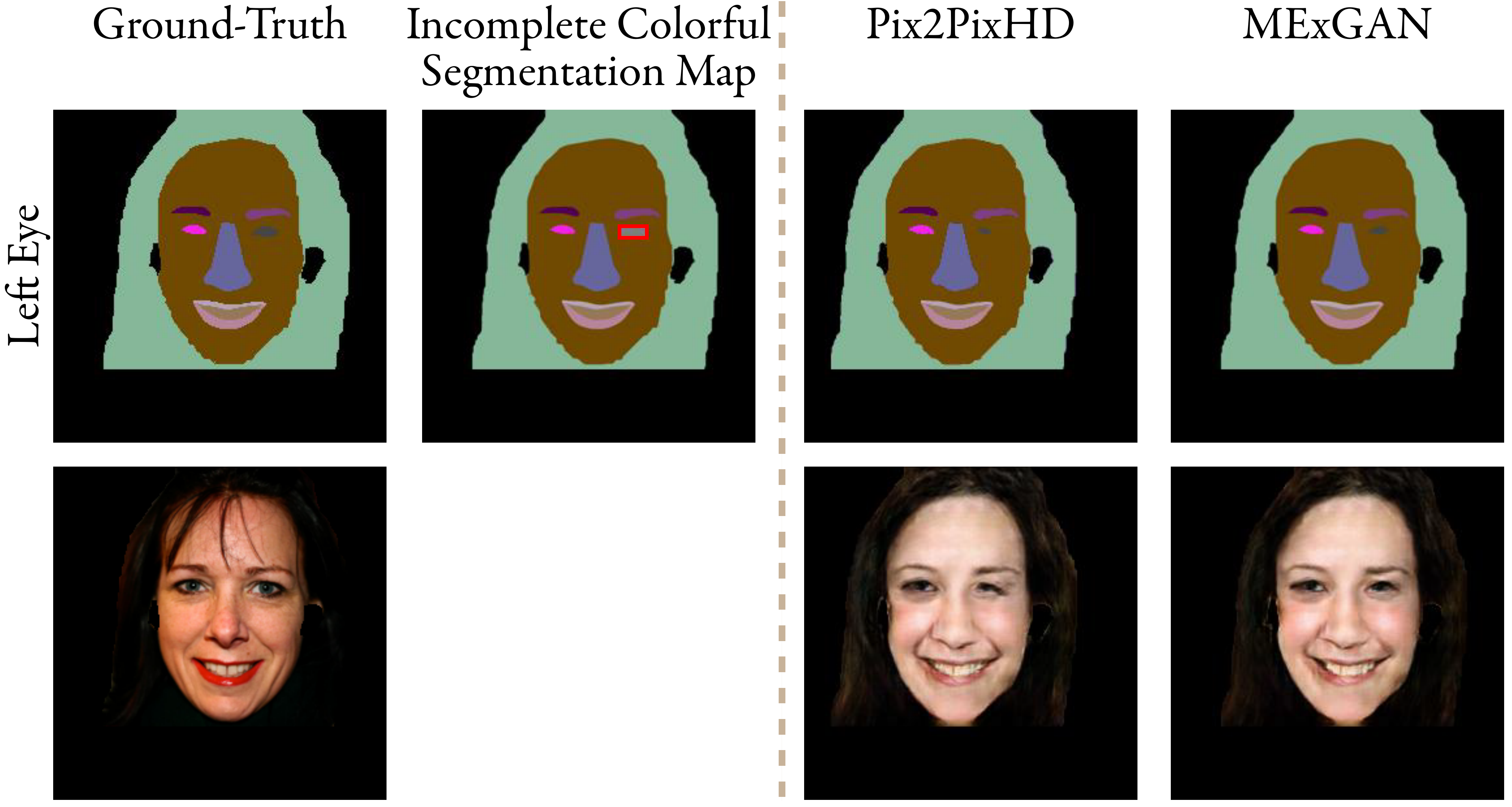}
\caption{Examples of manipulated segmentation maps and generated natural images on Helen Face. The images are arranged in the same way as Fig.~\ref{fig:Cityscapes}}
\label{fig:HelenFace}
\end{figure}

\begin{figure}[htb]
\centering
\includegraphics[width = 0.45\textwidth]{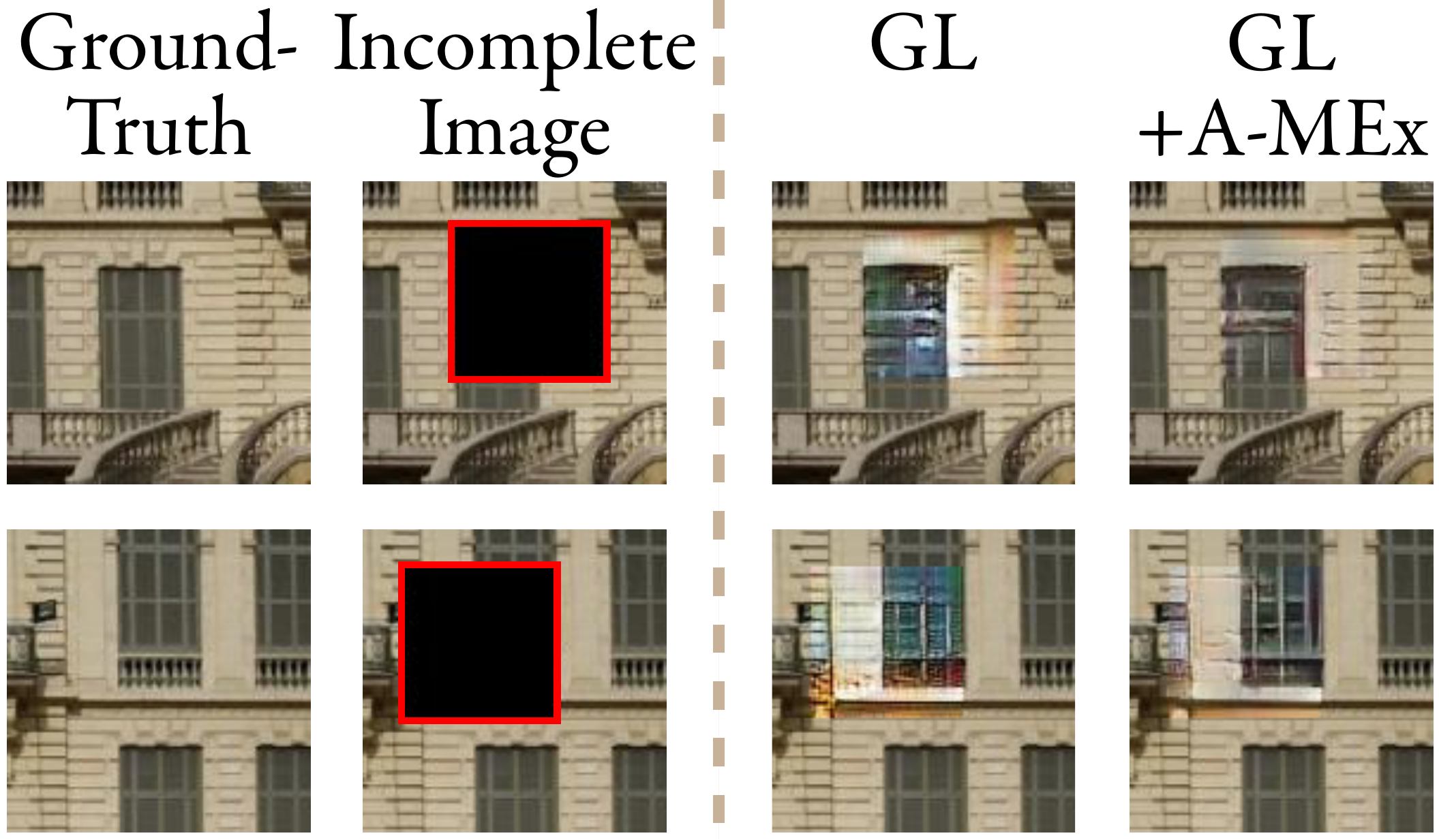}
\caption{Examples of image inpainting results on CMP Facade. The two left columns show ground-truth and incomplete image respectively. The third and fourth columns show the results from GL and GL+A-MEx respectively.}
\label{fig:Facade}
\end{figure}

\begin{figure}[htb]
\centering
\includegraphics[width = 0.45\textwidth]{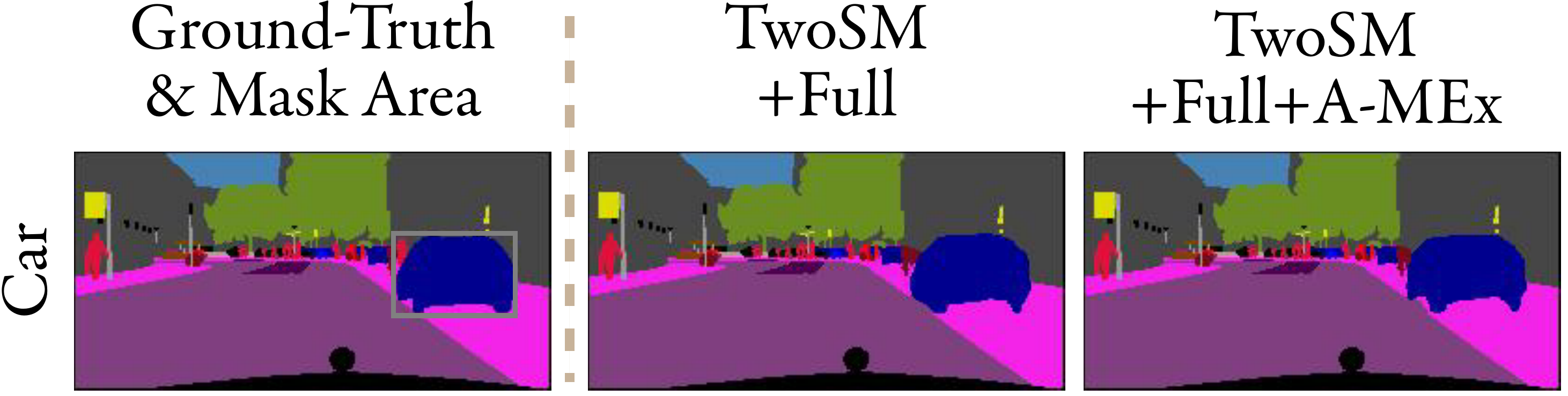}
\caption{Examples of manipulated segmentation maps on Cityscapes. The first column shows ground-truth and mask area (by grey box). The two right  columns show the results from TwoSM+Full, and TwoSM+Full+A-MEx respectively.}
\label{fig:NIPS18}
\end{figure}

\subsection{Experimental Results}
\subsubsection{Quantitative Results}
Tab.~\ref{table:pixHD},~\ref{table:TwoSM} and \ref{table:Mem} report quantitative results of our task by tIOU, hamm, and FID. Tab.~\ref{table:GL} reports quantitative results of natural image inpainting by SSIM, LI and FID. From these tables, we can conclude as below,

(1) \textbf{MEx loss is effective on segmentation map based on Pix2PixHD.} From Tab.~\ref{table:pixHD}, MEx loss improves each metric result in each dataset. For the manipulated segmentation map, the hamm has improved 24.45\% in NYU by MEx loss, and the tIOU has improved 32.71\% in Helen face by MEx loss. For the translated natural image, we also see the slight improvement in FID, which shows that our manipulated segmentation maps are closer to the ground-truth.  

(2) \textbf{MEx loss improves image inpainting over GL.} 
From Tab.~\ref{table:GL}, we see improvement in SSID, L1 and FID on CMP Facade. Especially, the FID of generated images decreases 13.13\%, which is a great improvement in image quality. 

(3) \textbf{Our improvement, replacing local observation with global observation, is effective.}  Tab.~\ref{table:TwoSM} shows TwoSM+Full obviously improves over TwoSM. For examples, the tIOU improves 17.28\% and hamm improves 13.22\%. These demonstrate that the removed context, which has been kept in the global observation, is meaningful to improve the performance of semantic editing on segmentation map.

(4) \textbf{MEx loss is not effective in TwoSM due to its double-stream structure. } From Tab.~\ref{table:TwoSM}, introducing MEx loss or A-MEx loss into TwoSM+Full leads to negligible decrease in tIOU and hamm, which means MEx loss is not effective in TwoSM. We apply MEx loss in TwoSM by adding it to the stream that generates background map, which includes both object shape and its surrounded context. The reason for negligible decrease might be that its two output maps cause noises for the shared embedding layers, if the object shape in foreground and object shape in background map do not overlap perfectly. In other word, the shared embedding layers of background map generation process with MEx loss are overwritten by foreground map generation process, which has no effect from MEx loss. 

(5) \textbf{A-MEx loss improves performance compared with MEx loss, while MEx loss saves more memory.} From Tab.~\ref{table:Mem}, MExGAN performs better in tIOU and hamm comapred with A-MExGAN. For example, A-MExGAN improves 2.12\% in tIOU compared with MExGAN. For the \textbf{GPU memory average consumption (Mem)}, the Mem for MExGAN is 2.42\% less than A-MExGAN. Though A-MExGAN has only $1$ MEx discriminator compared with MExGAN, which has $q$ MEx discriminators, the MEx areas for MExGAN have smaller size because of $Crop$ operation. The Mem shows that the $Crop$ operation saves more GPU memory compared with less MEx discriminators.

\subsubsection{Qualitative Results and Human Evaluations} The qualitative results are shown by  Fig.~\ref{fig:example1},~\ref{fig:Cityscapes},~\ref{fig:HelenFace},~\ref{fig:Facade},~\ref{fig:NIPS18} and HuEv.

\textbf{Image examples in semantic editing on segmentation map.} The Fig.~\ref{fig:example1} demonstrates the results our model adding or replacing an object on the segmentation map. Fig.~\ref{fig:Cityscapes} shows the manipulated segmentation maps and their translated natural images on the Cityscapes. From the Fig.~\ref{fig:Cityscapes}, MExGAN generates objects with diverse but reasonably looking shapes compared with ground-truth segmentation maps. Though the translated results might not reach state-of-the-art, the image translation is not our main task, which can be improved by other downstream modules. From the Fig.~\ref{fig:HelenFace}, we can see that the generated left eye on a face by MExGAN in a natural image appears more natural in comparison with that generated by Pix2PixHD. These results show the effectiveness of MEx loss in Pix2PixHD. As shown in Fig.~\ref{fig:NIPS18}, A-MEx loss generates more reasonable details in the left tires of the car, which shows that A-MEx loss is still effective on TwoSM+Full in some cases.

\textbf{Image examples in image inpainting.} Fig.~\ref{fig:Facade} shows examples of natural image inpainting on CMP Facade.  Fig.~\ref{fig:Facade} shows that the image quality , especially``seamless", is significantly boosted by A-MEx loss, which demonstrates the effect of A-MEx loss in natural image inpainting.

\textbf{Human Evaluation.} To measure the image quality more accurately, we conduct experiments for human evaluation, where the results are reported as HuEv in Tab.~\ref{table:pixHD},~\ref{table:TwoSM} and \ref{table:GL}. For HuEv (by 11 users) in Tab.~\ref{table:pixHD}, MExGAN generates better manipulated segmentation maps by having 12.59\% more users preferring MExGAN generation. For HuEv (by 12 users) in Tab.~\ref{table:GL}, an
improvement (11.55\%) is achieved by A-MEx loss. For HuEv (by 5 users) in Tab.~\ref{table:TwoSM},  a slight decrease (5.37\%) in MEx loss is observed, where the reason has been explained in (4) of Sec. ``Quantitative Results".

\begin{table}[htb]
\begin{center}
\begin{tabular}{l|lll}
\hline
\multirow{2}{*}{Methods} & \multicolumn{3}{c}{Cityscapes}\\
                         & tIOU &  hamm &HuEv \\
\hline
TwoSM                    & 0.7064    & 0.7443 & -  \\
\hline
TwoSM+Full            & \bf0.8285    &   \bf0.8427  &\bf0.5138 \\
TwoSM+Full+MEx             & 0.8242      & 0.8395  &-   \\
TwoSM+Full+A-MEx           & 0.8243    & 0.8387  &0.4862 \\
\hline
\end{tabular}
\caption{The tIOU, and hamm for ablation methods of TwoSM on the  Cityscapes. Specifically, we compare human evaluations on TwoSM+Full and TwoSM+Full+A-MEx.}
\label{table:TwoSM}
\end{center}
\end{table}

\begin{table}[htb]
\begin{center}
\begin{tabular}{l|llll}
\hline
\multirow{2}{*}{Methods} & \multicolumn{3}{c}{Cityscapes}\\
                         & SSIM & L1 &  FID &HuEv\\
\hline
GL                    & 0.3827   & 21.59    & 63.14  &0.4727 \\
\hline
GL+A-MEx            & \bf0.3912    & \bf20.22    &   \bf54.84 &\bf0.5273 \\
\hline
\end{tabular}
\caption{The SSIM, L1, FID and HuEv on CMP Facade for natural image inpainting.}
\label{table:GL}
\end{center}
\end{table}

\begin{table}[htb]
\begin{center}
\begin{tabular}{l|lllll}
\hline
\multirow{2}{*}{Methods} & \multicolumn{3}{c}{Cityscapes}\\
                          & tIOU &  hamm & Mem\\
\hline
MExGAN                   & 0.7493   & 0.7677 & \bf4.43GB \\
\hline
A-MExGAN            & \bf0.7652    & \bf0.7734  & 4.54GB \\
\hline
\end{tabular}
\caption{On the Cityscapes, the tIOU, hamm and Mem between MExGAN and A-MExGAN.}
\label{table:Mem}
\end{center}
\end{table}


\subsection{Discussion Based On Experimental Results}
\textbf{Ubiquity}: Though we propose a basic structure generator based on Pix2PixHD, the generator can be replaced with other models for other applications, where the replaced generator should not overwrite the output, such as natural image inpainting with GL.
\textbf{Limitation}: The MEx loss is not effective in a two-stream model, if we only add MEx loss to one stream, such as our usage of MEx loss in TwoSM. 



\section{Conclusion}
A framework to improve performance of semantic editing on segmentation map, called MExGAN, is proposed in this paper. To address misalignment issue, MExGAN applies a novel MEx loss implemented by adversarial losses on MEx areas.
To boost convenience and stability, 
we approximate MEx loss as A-MEx loss. Besides, MExGAN applies global observation to improve current data preparation, and can flexibly replace the basic structure generator in other applications. Extensive experiments on four datasets demonstrate effectiveness of global observation and MEx loss. 
In the future, we will generate diverse outputs for each condition.

\bibliography{main}

\subsection{Appendix}
\subsection{Organization of Technical Appendix}
In this appendix, we show the Technical Appendix of the paper ``Semantic Editing On Segmentation Map Via Multi-Expansion Loss". Firstly, we present additional quantitative and qualitative experiment results as well as more experiment setup, which provides further details about our experiments. Secondly, we clarify the items required in reproducibility checklist in details.

\subsection{Appendix of Experiments}
\subsubsection{Additional Experiment Setup}
Besides the experiment setup introduced in Sec. 4.1 of main script, we introduce more experiment setup as below.

\textbf{Editing categories} For NYU, we choose $7$ categories
(`cabinet', `chair', `floor', `table', `wall', `window', `picture') to test semantic editing on segmentation map. For Helen Face,  we choose $8$ editing categories
(`left brow', `right brow', `left eye', `right eye', `nose', `upper lip', `inner mouth', `lower lip'). The natural image inpainting on CMP Facade does not require semantics as input, so the editing categories are not necessary for it. 

\textbf{Image size} The training and generated image resolutions are $256\times128$, $192\times144$ and $256\times 256$ respectively for Cityscapes, NYU and Helen Face. As for the natural image inpainting task, we set $248\times242$ as the training and generated image resolutions.

\textbf{Excluding mechanism} We exclude the object bounding boxes that are too small to carry significant scene content. For Cityscapes and NYU datasets in MExGAN (Pix2PixHD + MEx loss), we set the size threshold as 0.02, which filters bounding boxes that are smaller than 2\% of the size of inputted images. The filter threshold for Helen Face is 0.0003, which is much smaller compared with 0.02, because the `left eye' and `right eye' are quite tiny areas compared with inputted images. As for TwoSM+Full+MEx and TwoSM+Full+A-MEx in Cityscapes, we set the filter threshold as 0.01. For the images that include no object bounding box satisfying the size threshold, we skip the images in both the training and testing processes.

\textbf{Parameter setting} Since we set $r=5$ in Eq. 4 for perceptual loss, then we have 5 chosen layers, the chosen layers are same to Pix2PixHD and the $\mathbf{w} = [1/32, 1/16, 1/8, 1/4, 1]$.

\begin{figure}[htb]
\centering
\includegraphics[width = 0.45\textwidth]{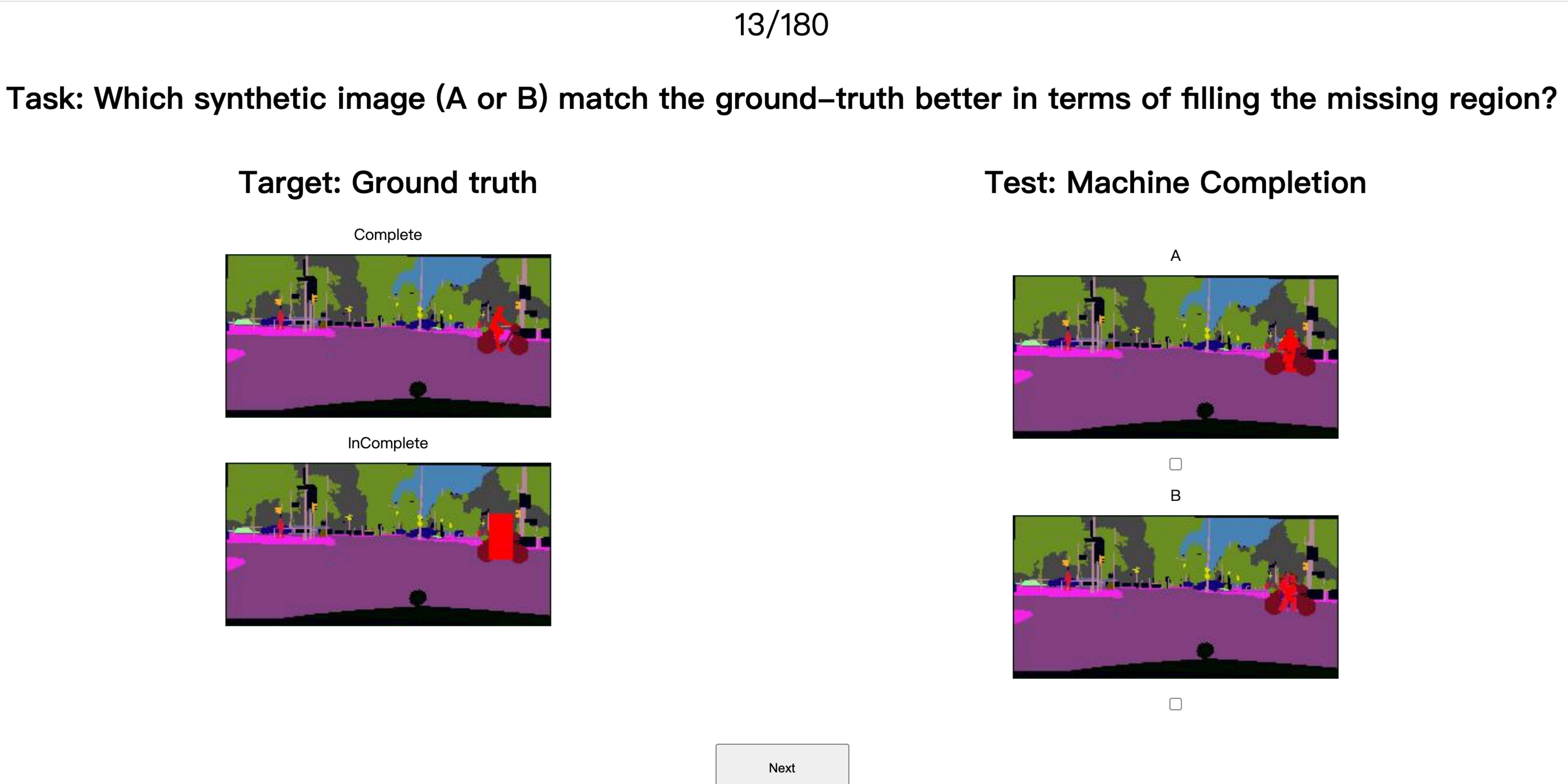}
\caption{A screenshot for the website used for user evaluation. In the top, we show the task question as ``Task: Which synthetic image (A or B) match the ground-truth better in terms of filling the missing region?" In the left part, the complete segmentation map, and incomplete segmentation map with object bounding box in certain color respecting to the target label are shown. For each turn, we provide random arranged results by choices A and B in the right part.}
\label{fig:user}
\end{figure}

\textbf{Details about user evaluation} We do not conduct user evaluation by Amazon Mechanical Turk (AMT), but by anonymous and unknown users from other organizations to finish the two-alternative forced choices experiments. The users are provided with a website for each experiments. A screen shot of the website is shown as Fig.~\ref{fig:user}, where the users know the complete segmentation map, and incomplete segmentation map, which shows the object bounding box in certain color respecting to the target label. For the right part of the website, the random arranged results (choices A and B) from the models are provided for comparison. In other words, the choices A and B in each turn randomly represent different model results.

\subsubsection{Additional Quantitative Results}
\begin{figure}[htb]
\centering
\includegraphics[width = 0.45\textwidth]{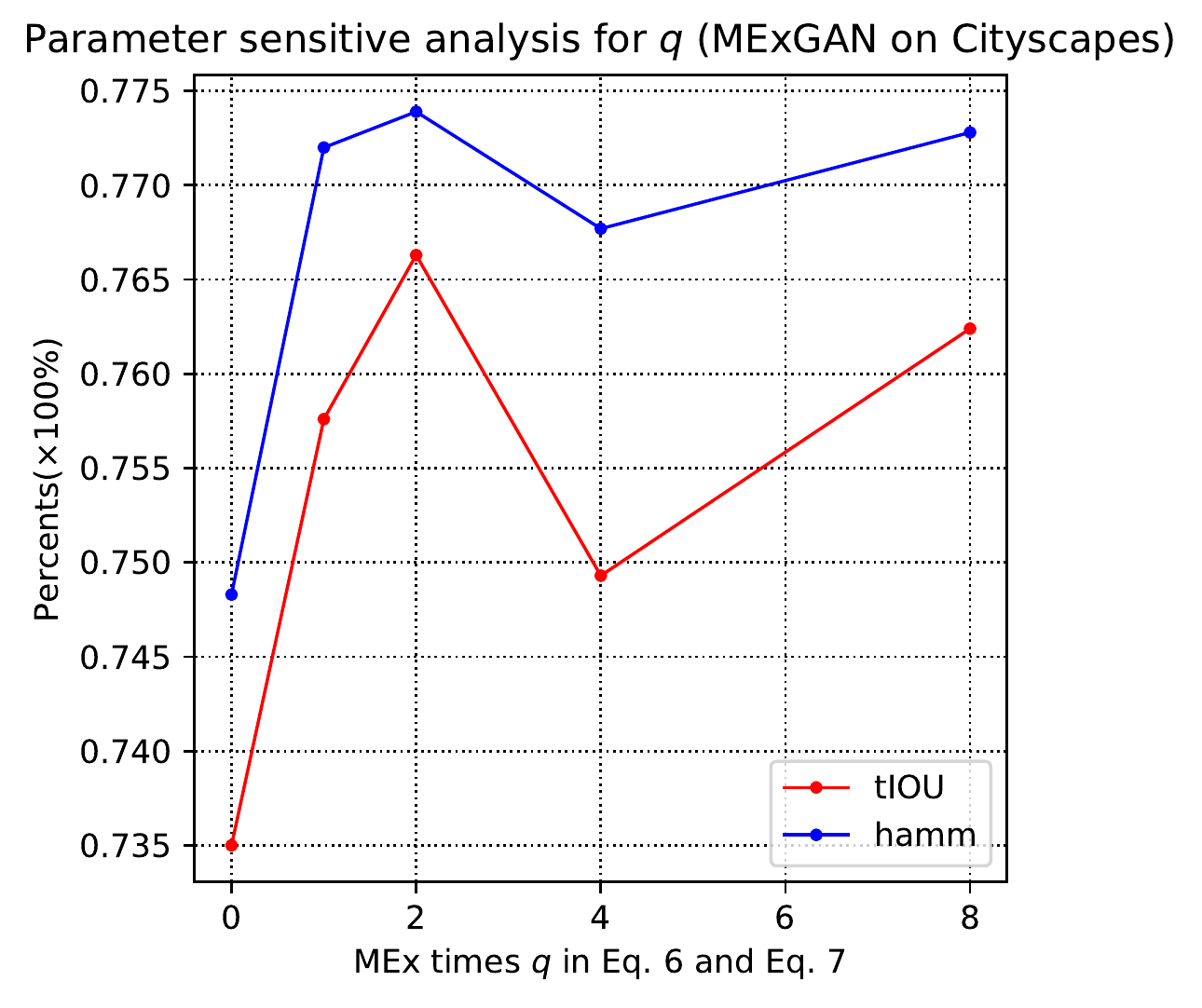}
\caption{Diagrams for parameter sensitive analysis on MEx times $q$. The tIOU and hamm are both obtained from MExGAN on Cityscapes. And the $q$ are set as 0, 1, 2, 4, 8. Moreover, we set $\alpha=\beta=5$ for all sensitive analysis experiments.}
\label{fig:sup_MEx_q}
\end{figure}

\textbf{Parameter sensitive analysis for $q$ by MExGAN on Cityscapes} Besides the quantitative results shown in Sec. 4.2 of main script, we also conduct experiments for the MEx times $q$ applied in Eq. 6 and Eq. 7 of main script. Concretely, we compare tIOU and hamm for different $q$ by MExGAN on Cityscapes, where the $q$ values are set as 0, 1, 2, 4, 8. Plus, the $\alpha$ and $\beta$ are both fixed as 5 in the analysis. 
The result of sensitive analysis is shown in Fig.~\ref{fig:sup_MEx_q}. From the figure, we can conclude as follows,

(1) The tIOU and hamm in $q=0$ is much lower by obvious difference compared with those in $q>0$. Since $q=0$ is equivalent to GL, this shows our proposed multiple mid-level views are effectively complementary views for global and local views.

(2) The tIOU and hamm are both fluctuated by the increase of $q$. In our setting, when $q=2$, we have the highest tIOU and hamm, while the tIOU and hamm in $q=4$ decrease around 1\% compared with those in $q=2$. However, when $q=8$, the tIOU and hamm increase obviously compared those in $q=4$. The fluctuation shows that tIOU and hamm are sensitive to the $q$.

Though the tIOU and hamm in $q=2$ achieve more competitive results, we still report those in $q=4$, which leads to the lowest performance when $q>1$, in the main script. The reason is that we do not have validation dataset, which is same as Pix2PixHD~\protect\cite{wang2018high} and TwoSM~\protect\cite{hong2018learning}, and it is unfair to choose the parameters resulting in the best performance in test dataset. Thus, we choose $q=4$ on Cityscapes, which still beats performance in $q=0$ obviously. 

\subsubsection{Additional Qualitative Results}
\begin{figure*}[htb]
\centering
\includegraphics[width = 0.95\textwidth]{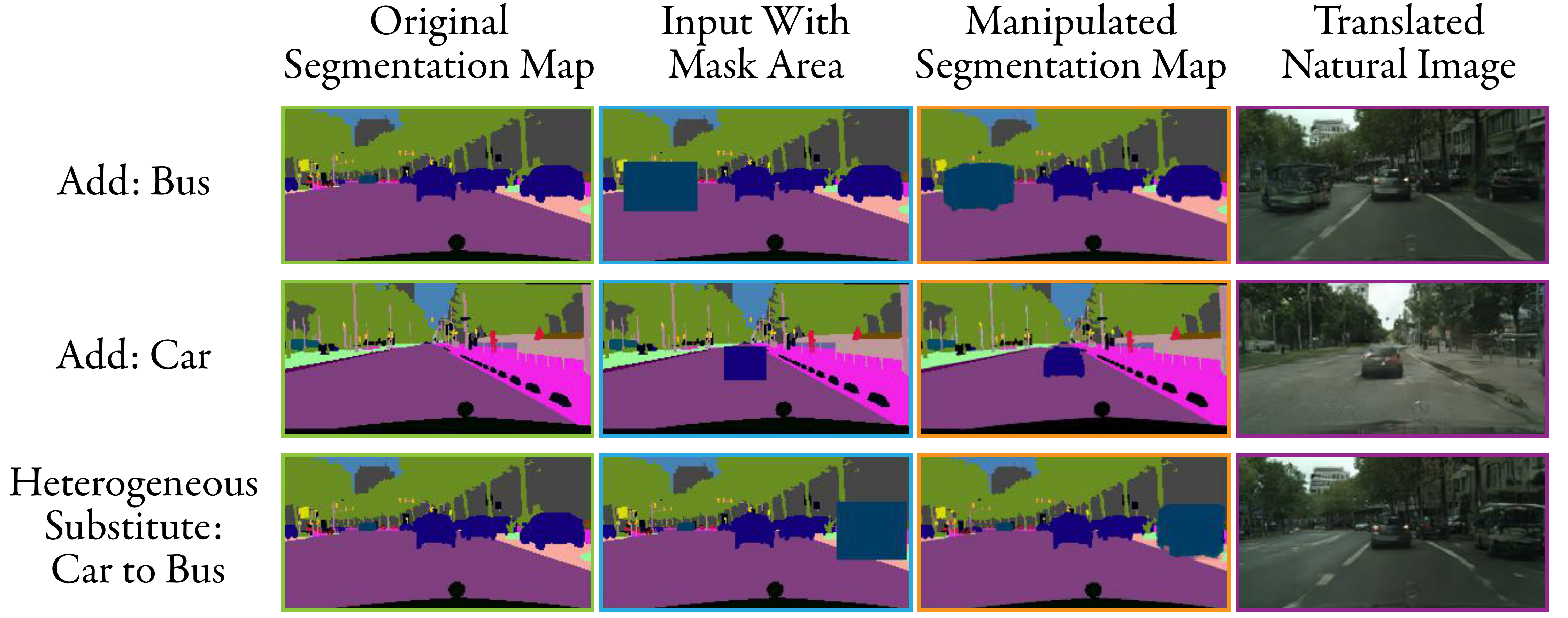}
\caption{Examples of image editing using MExGAN. Our method directly manipulates scene contents from the segmentation maps by selecting a mask area and a target label, which informs the model how to synthesize manipulated segmentation maps and generate natural images. From left to right: 1) the complete segmentation maps in color, 2) the incomplete segmentation maps with mask area painted with color respecting to the target label, 3) the manipulated segmentation maps by MExGAN, 4) the translated natural images from the manipulated segmentation maps.}
\label{fig:sup_1}
\end{figure*}

\begin{figure*}[htb]
\centering
\includegraphics[width = 0.95\textwidth]{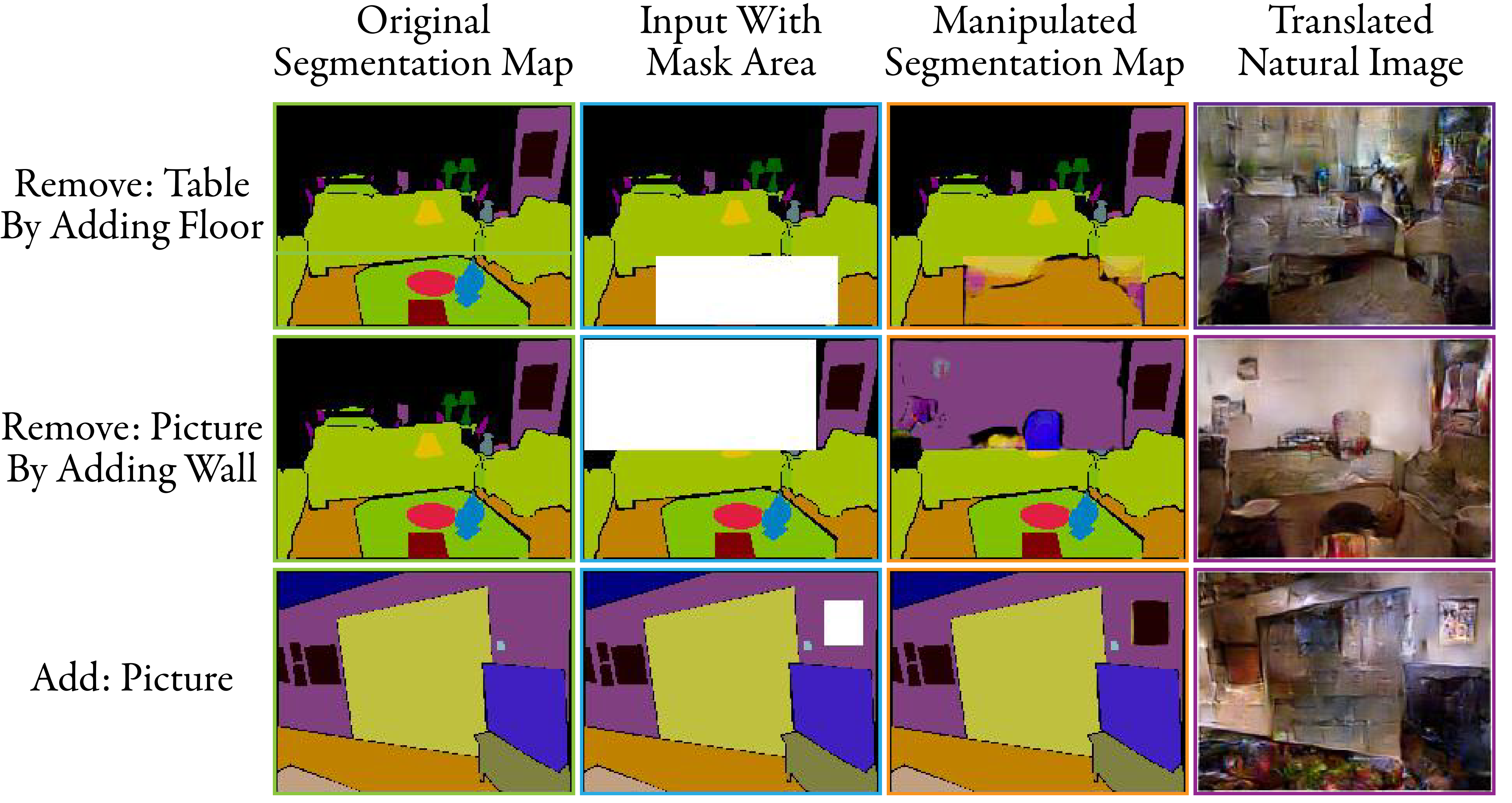}
\caption{Examples of image editing by setting mask areas and target labels on NYU dataset. The white rectangles represent mask areas. The images are arranged in the same way as Fig.~\ref{fig:sup_1}}
\label{fig:sup_2}
\end{figure*}

\begin{figure*}[htb]
\centering
\includegraphics[width = 0.95\textwidth]{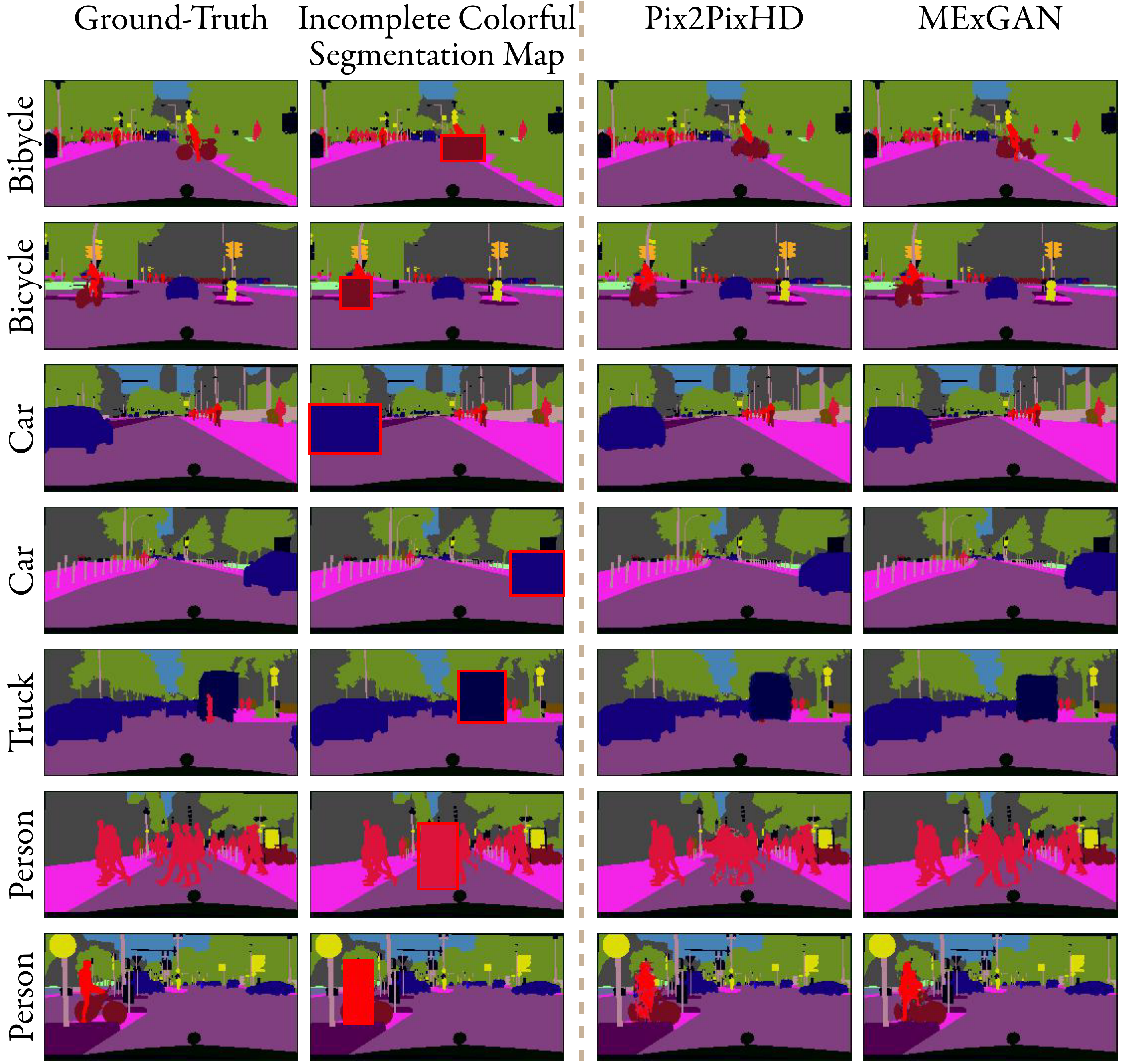}
\caption{Examples of manipulated segmentation maps of Pix2PixHD and MExGAN on the Cityscapes. The ground-truth segmentation maps, and the incomplete color segmentation maps are shown on the left. The right two columns show the manipulated segmentation maps of Pix2PixHD and MExGAN respectively. The leftmost vertical texts are their target labels.}
\label{fig:sup_Pix2PixHD}
\end{figure*}

\begin{figure*}[htb]
\centering
\includegraphics[width = 0.95\textwidth]{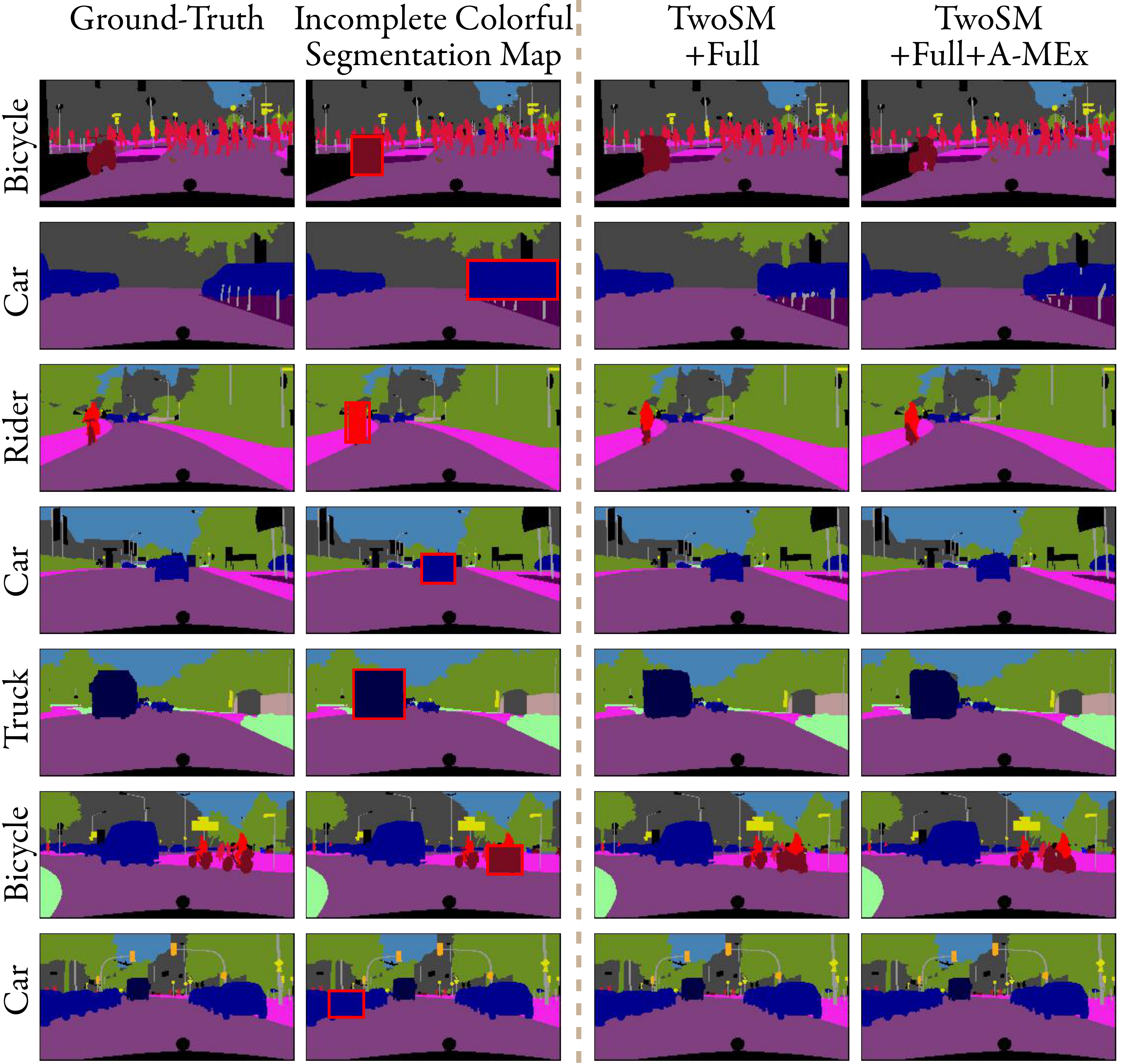}
\caption{Examples of manipulated segmentation maps of TwoSM+Full and TwoSM+Full+A-MEx on the Cityscapes. The ground-truth segmentation maps, and the incomplete color segmentation maps are shown on the left. The right two columns show the manipulated segmentation maps of TwoSM+Full and TwoSM+Full+A-MEx respectively. The leftmost vertical texts are their target labels.}
\label{fig:sup_3}
\end{figure*}

\begin{figure*}[htb]
\centering
\includegraphics[width = 0.99\textwidth]{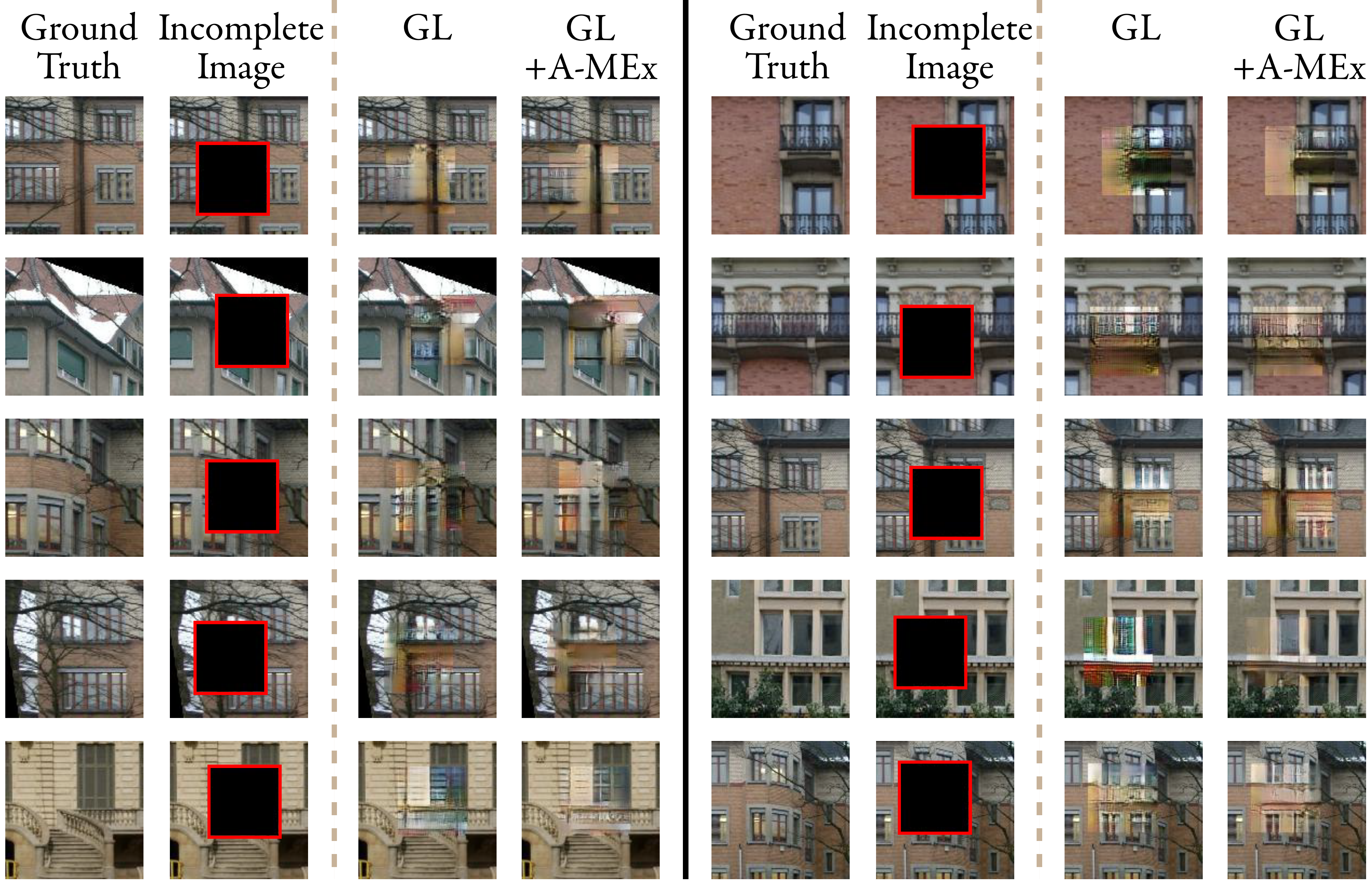}
\caption{Examples of image inpainting results on CMP Facade. The figure can be divided into two parts, left four columns and right four columns. The left two columns show ground-truth and incomplete image with mask area respectively. The left third and fourth columns show the results from GL and GL+A-MEx respectively. The right part has same arrangement as the left part.}
\label{fig:sup_4}
\end{figure*}

The Fig.~\ref{fig:sup_1}, Fig.~\ref{fig:sup_2}, Fig.~\ref{fig:sup_Pix2PixHD} and Fig.~\ref{fig:sup_3} show additional qualitative results of semantic editing on segmentation map. In the Fig.~\ref{fig:sup_4}, we show extra qualitative results of image inpainting.   

\textbf{Application of semantic editing on segmentation map with downstream task as image translation} In the Fig.~\ref{fig:sup_1} and Fig.~\ref{fig:sup_2}, which are results from Cityscapes and NYU respectively, we show the results of semantic editing on segmentation maps and the translated natural images from the manipulated segmentation maps. From the two figures, one can see that though we only add the objects by semantics, we can also implement removal and substitution by the ``add" operation. For examples, we can replace a car with a bus by masking an area including the car and giving a target label ``bus", shown in the third row in Fig.~\ref{fig:sup_1}. As for the removal, we can add the background in a mask area, such as adding floor in a mask area, shown in the first row of Fig.~\ref{fig:sup_2}.

\textbf{Additional manipulated segmentation maps of Pix2PixHD affected by MEx loss} Fig.~\ref{fig:sup_Pix2PixHD} shows additional qualitative results, which demonstrates effectiveness of MEx loss on Pix2PixHD. For examples, its second row shows MEx loss can help generate more competitive bicycle. The person in the last row is generated more reasonably by MEx loss.

\textbf{Additional manipulated segmentation maps of TwoSM affected by MEx loss} Due to the two-stream structure of TwoSM, our proposed MEx loss is not effective from view of quantitative results. However, Fig.~\ref{fig:sup_3} shows additional cases where TwoSM+Full+A-MEx has better qualitative performance than TwoSM+Full in different semantics. For examples, its first row shows better generation in bicycle from TwoSM+Full+A-MEx, because of its more pronounced contour of the bicycle. Its last row shows better car, because the car generated by TwoSM+Full+A-MEx has more clear tires compared with the one in TwoSM+Full.

\textbf{Additional natural images of GL affected by MEx loss} From the Fig.~\ref{fig:sup_4}, GL+A-MEx shows obviously improvement in image quality by A-MEx loss. For examples, the first row in left part shows more details about the generated window, which is more consistent with its context. The last row in left part shows a very seamless results by GL+A-MEx. In addition, we also observe that the results of GL+A-MEx have less noise compared with those of GL. The reason might be that the MEx loss provides more multiple mid-level views, besides global and local views, so that it is more robust compared with GL.

\subsection{Reproducibility Checklist}
We emphasize on the reproducibility checklist in terms of two parts, datasets and computational experiments. Besides, we have attached our data samples and project code with its ReadMe in the data\_code appendix. Plus, we will release our project code with ReadMe to GitHub, if our paper is accepted.

\subsubsection{Dataset Reproducibility}
We do not have any novel datasets in our experiments. And we show the reproducibility of all datasets applied in our experiments in this section. Concretely, we introduce more information about the four commonly-used datasets, Cityscapes, NYU V2, rectified Helen Face and CMP Facade datasets.

\textbf{Cityscapes} The dataset can be downloaded from ``https://www.cityscapes-dataset.com/downloads/", where we download ``gtFine\_trainvaltest.zip (241MB)" and ``leftImg8bit\_trainvaltest.zip (11GB)", which provide segmentation maps and natural images respectively.

\textbf{NYU V2} It is downloaded from ``https://cs.nyu.edu/~silberman/datasets/nyu\_depth\_v2.html", where we only download ``Labeled dataset (2.8 GB)", which provides us both segmentation maps and natural images. However, different from Cityscapes, which provides segmentation maps and natural images by various images, NYU V2 saves all data by a ``.mat" file.

\textbf{Rectified Helen Face} We download it from ``https://github.com/JPlin/Relabeled-HELEN-Dataset". Instead of applying originally natural images for training, same to Pix2PixHD, we mask the background information of the originally natural images by their respective segmentation maps.

\textbf{CMP Facade} It is provided by ``http://cmp.felk.cvut.cz/$\sim$tylecr1/facade/", where we download both the base dataset and extended dataset. 

As for the division of training dataset and testing dataset, the first three datasets provide default training and testing datasets, which are applied in our experiments. Since the CMP Facade does not provide default training and testing datasets, we follow the setting in ``https://github.com/sevmardi/gl\_Image\_Inapinting\_pytorch".

\subsubsection{Computational Experiment Reproducibility}
We further answer reproducibility questions about computational experiments in this section.  

\textbf{Randomness} Since in the both training and testing processes, our work randomly mask an object bounding box to prepare data. In the training process, we do not set a random seed, because we wish the randomness between different epochs is different, so that we can mask more object bounding boxes for training. However, in the testing process, we set a fixed random seed and only run 1 epoch, because we wish all comparison experiments mask the same object bounding box for each image. We set random seed as ``679" for all datasets. However, we find that even though the TwoSM and Pix2PixHD have been set same random seed, their masked object bounding boxes are still different, this is one of the reasons why we do not compare them in one table.

\textbf{Computing infrastructure} We do all experiments on two GPUs, which both are GTX 1080Ti. The RAM in our machine is 64 GB. The key versions of relevant software libraries are available by GitHub of the basic structure generators. For MExGAN, which has basic structure generator as Pix2PixHD, we can find the key versions from ``https://github.com/NVIDIA/pix2pixHD". For TwoSM+Full+MEx, they are available at ``https://github.com/xcyan/neurips18\_hierchical\_image\_mani\\pulation". As for GL+A-MEx, we can find them from ``https://github.com/sevmardi/gl\_Image\_Inapinting\_pytorch".

\textbf{Motivation choosing the metrics} Though we have explained calculation of the metrics in Sec. 4.1 of main script, we explain the reasons for choosing these metrics as below. 

We have four evaluation metrics for semantic editing on segmentation map. We choose tIOU because it can scale the overlapping degree between the generated target object and ground-truth target object. However, if we only evaluate by tIOU, then the bias situations will be that the generated target object is extremely huge and the tIOU is very high. Thus, we choose hamm, which scale the correctness for each pixel in the mask area. A better generation should have both higher tIOU and hamm. Since the downstream task in our work is image generation, we assume the better translated natural images we have, the better manipulated segmentation maps we generate. Thus, we apply FID to scale the image quality of translated natural images. 

Besides, we have two extra evaluation metrics for natural image inpainting. We choose SSIM and L1, because these two both scale the difference between the original images and the generation in statistics-level and pixel-level respectively. The less difference we have, the better image generation we have. Plus, FID as a common method to scale the image quality with a pretrained model, which shows the difference between distribution of generated images and distribution of real images. The smaller FID we have, the more natural images we generate.

Considering the generation is a subjective task and quantitative results might not represent human evaluations, we also have users to evaluate the results.

\textbf{Number of algorithm runs} The number of algorithm runs used to compute each reported result is 1, because we have set random seed in the testing process. As a result, the mask areas are fixed. Then, we also observe that the quantitative results will keep unchanged in different trials, if we keep random seed unchanged.

\textbf{Number and range of values tried per (hyper-)parameter} We try the 5 different values for $q$ by MExGAN on Cityscapes, which is reported in Sec 2.2 in Appendix. The range is concrete 5 values (0, 1, 2, 4, 8). For $\alpha$ and $\beta$, we do not conduct parameter sensitive analysis, but this can be done, if this is necessary or required. Besides, we sincerely 
apologize for making the wrong selection for this question, the correct answer should be "partial". 

\textbf{All final (hyper-)parameters used} We have given the final parameters in the Sec. 4.1 of main script and Sec. 2.1 in the appendix.

\end{document}